\documentclass[preprint,10pt,3p]{elsarticle}
\usepackage{multicol}
\usepackage{graphicx}%
\usepackage{multirow}%
\usepackage{amsmath,amssymb,amsfonts}%
\usepackage{textcomp,mathcomp}%
\usepackage{amsthm}%
\usepackage{mathrsfs}%
\usepackage[title]{appendix}%
\usepackage{xcolor}%
\usepackage{manyfoot}%
\usepackage{booktabs}%
\usepackage{algorithm}%
\usepackage{algorithmicx}%
\usepackage{algpseudocode}%
\usepackage{longtable} %
\usepackage{adjustbox}
\usepackage{listings}%
\usepackage{url}
\usepackage{hyperref}
\usepackage{booktabs}
\usepackage{array}
\usepackage{comment}%
\usepackage{adjustbox}%
\usepackage{graphicx}%
\usepackage{caption}
\usepackage{subcaption}
\usepackage{placeins}

\newcommand{\sinergym}{\textsc{Sinergym}}
\newcommand{\datacenter}{\textsc{2ZoneDataCenterHVAC}}

\definecolor{dkgreen}{rgb}{0,0.6,0}
\definecolor{gray}{rgb}{0.5,0.5,0.5}
\definecolor{mauve}{rgb}{0.58,0,0.82}

\newcommand{\added}[1]{1}

\hypersetup{
    colorlinks=true
}

\usepackage{lineno}

\journal{Energy and Buildings}

\graphicspath{{figures/}}

\begin{document}

\begin{frontmatter}

\title{\sinergym{} – A virtual testbed for building energy optimization with Reinforcement Learning}


\author[1]{Alejandro Campoy-Nieves\corref{cor1}}\ead{alejandroac79@correo.ugr.es}
\author{Antonio Manjavacas}\ead{manjavacas@ugr.es}
\author{Javier Jiménez-Raboso}\ead{jajimer@correo.ugr.es}
\author{Miguel Molina-Solana}\ead{miguelmolina@ugr.es}
\author{Juan Gómez-Romero}\ead{jgomez@decsai.ugr.es}

\cortext[cor1]{Corresponding author}

\affiliation{organization={Department of Computer Science and Artificial Intelligence},
            addressline={Universidad de Granada}, 
            city={Granada},
            postcode={18071}, 
            country={Spain}}

\begin{abstract}
Simulation has become a crucial tool for Building Energy Optimization (BEO) as it enables the evaluation of different design and control strategies at a low cost. Machine Learning (ML) algorithms can leverage large-scale simulations to learn optimal control from vast amounts of data without supervision, particularly under the Reinforcement Learning (RL) paradigm. Unfortunately, the lack of open and standardized tools has hindered the widespread application of ML and RL to BEO. To address this issue, this paper presents \sinergym{}, an open-source Python-based virtual testbed for large-scale building simulation, data collection, continuous control, and experiment monitoring. \sinergym{} provides a consistent interface for training and running controllers, predefined benchmarks, experiment visualization and replication support, and comprehensive documentation in a ready-to-use software library. This paper 1) highlights the main features of \sinergym{} in comparison to other existing frameworks, 2) describes its basic usage, and 3) demonstrates its applicability for RL-based BEO through several representative examples. By integrating simulation, data, and control, \sinergym{} supports the development of intelligent, data-driven applications for more efficient and responsive building operations, aligning with the objectives of digital twin technology.
\end{abstract}


\end{frontmatter}

\section{Introduction}\label{sec:introduction}
Buildings account for about 30\% of global energy consumption and are responsible for 17\% and 10\% of global direct and indirect global CO$_2$ emissions, respectively, according to the International Energy Agency\footnote{\url{https://www.iea.org/reports/buildings}} and the United Nations Environment Program\footnote{\url{https://www.unep.org/resources/report/2021-global-status-report-buildings-and-construction}}. The primary source of this consumption comes from Heating, Ventilation and Air Conditioning (HVAC) systems, representing more than 50\% of their associated energy demand in developed countries \cite{perez2008}. This consumption is potentially affected by the lack of precise control over these systems, whose proper and efficient operation is essential to ensure energy savings \cite{wang2020a,mawson2021,gholamzadehmir2020}.

Building Energy Optimization (BEO) aims to reduce energy consumption in buildings while maintaining occupant comfort. BEO methods mainly rely on simulated environments to test, compare and improve energy control strategies that can then be implemented in real buildings. By reducing the risks and costs of testing new strategies, simulation serves as a key enabler of digital twin technology, where virtual models support real-time, intelligent building operations. In this context, Machine Learning (ML) can leverage simulators to automatically learn optimal control strategies from massive data, being often able to outperform traditional methods \cite{CCAI}. 

Conventional energy analysis systems, such as eQuest\footnote{\url{https://www.doe2.com/equest/}} and Carrier’s Building System Optimizer\footnote{\url{https://www.carrier.com/commercial/en/us/software/hvac-system-design/building-system-optimizer}}, are useful for simulating energy consumption and configuring variables of HVAC systems during the design phase. However, these tools rely on static modeling, wherein operational parameters and control schedules remain fixed throughout each simulation. While general-purpose simulators like EnergyPlus\footnote{\url{https://energyplus.net}} and Modelica\footnote{\url{https://modelica.org}} offer more flexibility for performance analysis and equipment sizing, they are primarily designed for offline assessing and similarly lack the feedback loops required for continuous control. Specifically, they do not enable real-time adjustments of control strategies or assess the effectiveness of such actions as conditions evolve \cite{findeis2022}. This highlights the need for tools that integrate dynamic control features with energy simulation, as demonstrated by BOPTEST for Model Predictive Control in conjunction with Modelica \cite{blum2021}.

In the last years, Reinforcement Learning (RL) algorithms have emerged as the predominant approach to data-driven BEO. RL is a paradigm within ML based on learning an optimal behaviour policy through trial and error. This policy must be able to establish a proper mapping between states and actions in order to maximise a reward signal \cite{sutton2018}. Deep RL (DRL) extends RL by using neural networks to approximate value functions or policies \cite{hao2020}. RL and DRL have shown better performance than control alternatives such as Rule-Based Controllers (RBC) and Model Predictive Control (MPC) in certain scenarios \cite{wei2017,mason2019,zhang2018,vazquez2019,brandi2020,azuatalam2020,yu2021,perera2021,fu2022,deng2022,mahbod2022}, although they face several challenges derived from their data inefficiency \cite{NAGY2023110435}. Moreover, the application of RL-based BEO generally requires prior training in a simulated environment, which involves several challenges:

\begin{enumerate}
    \item \textbf{The setup of the building simulation software for different use cases} \cite{findeis2022}. As mentioned, simulation engines are not adequate for dynamic and continuous control, and particularly, for the observation-action-reward cycle of RL. Furthermore, configuring these simulators may be challenging because they have heterogeneous communication interfaces and data structures.

    \item \textbf{The execution and management of a large number of simulations} \cite{mason2019}. A large number of simulations are needed to train RL models adapted to different scenarios, building dynamics, and control strategies, requiring frequent restarts and warm-ups. In addition, these simulations must not only be executed efficiently but also monitored, thus requiring a robust back-end engine and flexible interfaces.

    \item \textbf{The reproducibility of the results} \cite{yu2021}. Many RL approaches to BEO address similar problems under different, partially defined configurations. Moreover, RL algorithms can exhibit stochastic behaviour, which makes them difficult to compare.

    \item \textbf{The comparison between controllers' performance} \cite{vazquez2019}. The absence of an established benchmark for RL-based BEO, in contrast to other ML areas ---e.g., image classification, translation, etc.--- prevents the standardized assessment of control solutions. Furthermore, as previously mentioned, the different parameters associated with each optimization technique require precise definitions and public disclosure to ensure transparent comparison.
   
\end{enumerate}

In this paper, we present \sinergym{}\footnote{\url{https://github.com/ugr-sail/sinergym}}, an open-source virtual testbed for RL-based BEO that addresses and solves the aforementioned issues. \sinergym{} offers a scalable, customizable, and robust platform for configuring, running, tracking, and collecting data from massive building simulations, along with specific features for training DRL agents. The software facilitates the inclusion of additional building models, and promotes the fair comparison between algorithms and the resulting control strategies. By leveraging EnergyPlus for simulation, \sinergym{} empowers researchers and practitioners to develop, test, and optimize intelligent control strategies that can transition from virtual environments to real-world building operations. Its unique features have been exploited for comparing state-of-the-art DRL algorithms \cite{wang2023,manjavacas_experimental_2024}, developing new RL approaches \cite{dmitrewski2022,Clue2023}, creating ML datasets \cite{liu2022}, learning digital twins \cite{naug2023}, explaining control strategies \cite{Jimenez2023}, applying federated learning \cite{hagstrom_employing_2024}, and testing controllers before deploying \cite{wolfle2023, jang2023}, among others.

The rest of the paper is structured as follows: Section~\ref{sec:relatedwork} reviews existing frameworks for simulation-based BEO and compares their main features. Section~\ref{sec:design} presents the rationale behind the design of \sinergym{}, Section~\ref{sec:functionalities} details its main functionalities, and Section~\ref{sec:examples} provides some representative usage examples. Finally, Section~\ref{sec:conclusions} summarizes the work and discusses the next steps for the tool.

\section{Related Work}
\label{sec:relatedwork}

\subsection{Interplay of RL algorithms and building simulators}
\label{subsec:drl_beo}

Simulation-based BEO involves the interaction between a controller and a building simulation engine, such as EnergyPlus and Modelica. Specifically, RL agents interact with the simulated building ---the \textit{environment}, in RL terminology--- by sending control actions to different building subsystems. As a result, the building transitions to a new state, computed by the simulation engine, resulting in a partial observation of the building state and a reward signal, representing the goodness of the transition.

The Gymnasium API\footnote{\url{https://gymnasium.farama.org/index.html}} \cite{gymnasium2024} (formerly OpenAI Gym \cite{brockman2016}) has become the standard interface for step-by-step interaction between RL agents and simulation environments. It defines three fundamental functions for this interaction: \texttt{init}, \texttt{step}, and \texttt{reset}, along with defined data structures for action and observation spaces. Any RL agent can easily interact with an environment that implements these functions, leading to the widespread adoption of standard DRL libraries, such as StableBaselines3 \cite{raffin2021} or RLlib \cite{liang2018}. Since the Gymnasium interface aligns with any control loop, it can be directly used by different types of controllers. 
 
Nevertheless, integrating the Gymnasium API with a building simulator is not straightforward \cite{wang2020b,biemann2021}. One reason is that building simulation engines are not intended for continuous and extensive interaction, but for running batch simulations during long periods. Furthermore, the execution of building simulation software tends to be costly due to the complex equations of thermodynamic physics laws that have to be solved in real time. Consequently, several frameworks aimed at BEO via building simulations have emerged in recent years. We analyse and compare them in the following subsection.

\subsection{RL-oriented frameworks for BEO}
\label{subsec:frameworks}

All the frameworks presented below aim at simplifying the interaction between custom controllers and back-end building simulators, preferably through the Gym/Gymnasium interface, which abstracts lower-level implementation details. The interaction with the building simulation is also continuous, so control signals are sent with user-defined temporal granularity, while the resulting states can be observed after each control action is performed.

Besides \sinergym{}, we identify three widely used frameworks for RL-based building energy optimization: RL Testbed for EnergyPlus \cite{moriyama2018}, BOPTEST-Gym \cite{arroyo2021}, and Energym \cite{scharnhorst2021}. In this section, we focus our comparison on these tools, excluding other frameworks that appear to be no longer actively developed, such as Gym-Eplus \cite{wei2017}, ModelicaGym \cite{lukianykhin2020}, Tropical Precooling Environment \cite{wolfle2020}, COmprehensive Building Simulator (COBS) \cite{zhang2020b}, and RL-EmsPy\footnote{\url{https://github.com/mechyai/RL-EmsPy}}. Additionally, we discuss the role of CityLearn \cite{nweye2024}, which targets building energy coordination and demand response across cities. We exclude tools like GridLearn \cite{pigott2021} and Grid2Op \cite{marot2021}, as they focus on energy distribution grids rather than building-specific applications.

\autoref{table:frameworks1} and \autoref{table:frameworks2} present a summary of the main features of the studied frameworks. They all include a Python layer over the simulation software to abstract the lower-level implementation details and enable continuous interaction. They rely either on Modelica or EnergyPlus as back-end simulators, with Energym being the only one including both of them. All but Energym are compatible with the Gym interface. However, only \sinergym{} and BOPTEST-Gym are fully compatible with the latest versions of the underlying simulation engines and the Gymnasium interface, which is currently the default standard in the RL domain.

\begin{table}[htpb]
\caption{Comparison between \sinergym{} and alternative building control frameworks -- Part 1}
\label{table:frameworks1}
\centering
\begin{tabular}{llllll}
\toprule
\textbf{Framework}                                                               & \textbf{Simulator}                                                                    & \textbf{Middleware}                                                              & \textbf{API}                                                 & \textbf{Buildings} & \textbf{Envs} \\ 
\midrule
\textbf{\begin{tabular}[c]{@{}l@{}}RL TestBed \end{tabular}} & \begin{tabular}[c]{@{}l@{}}EnergyPlus v9.5.0\end{tabular}                          & \begin{tabular}[c]{@{}l@{}}Custom patch \\ in simulator\end{tabular}             & \begin{tabular}[c]{@{}l@{}}Gym \\ v0.15.7\end{tabular}       & 1               & 3             \\ \hline
\textbf{Energym}                                                                 & \begin{tabular}[c]{@{}l@{}}EnergyPlus v9.5.0 \\ and Modelica v2.14\end{tabular} & \begin{tabular}[c]{@{}l@{}}Functional Mock-up \\ Interface (FMI)\end{tabular} & No                                                           & 7               & 14            \\ \hline
\textbf{\begin{tabular}[c]{@{}l@{}}BOPTEST-Gym\end{tabular}}                  & \begin{tabular}[c]{@{}l@{}}Modelica v4.0.0\end{tabular}                            & \begin{tabular}[c]{@{}l@{}}HTTP REST API \\ (based on FMI)\end{tabular}          & \begin{tabular}[c]{@{}l@{}}Gymnasium \\ v0.28.1\end{tabular} & 7               & 7             \\ \hline
\textbf{\sinergym{}}                                                                & \begin{tabular}[c]{@{}l@{}}EnergyPlus v24.1.0\end{tabular}                         & \begin{tabular}[c]{@{}l@{}}EnergyPlus \\ Python API v0.2\end{tabular}         & \begin{tabular}[c]{@{}l@{}}Gymnasium \\ v0.29.1\end{tabular} & 4               & 87            \\
\hline
\end{tabular}
\subcaption*{\textbf{Simulator}: Building simulation engine (EnergyPlus, Modelica or both). \textbf{Middleware}: Communication interface between Gym/Gymnasium and the simulator.
\textbf{API}: Is the tool compliant with the Gym/Gymnasium interface? \textbf{Buildings}: Number of different buildings included.
\textbf{Envs}: Number of predefined environments (buildings + configuration parameters)}
\end{table}

\begin{table}[htbp]
\caption{Comparison between \sinergym{} and alternative building control frameworks -- Part 2}
\label{table:frameworks2}
\renewcommand{\arraystretch}{1.5} 
\centering
\begin{tabular}{llllll}
\toprule
\textbf{Framework}                                                               & \textbf{WeatherSet} & \textbf{WeatherVar} & \textbf{Actions} & \textbf{DynamicSp} & \textbf{CustomRw} \\
\midrule
\textbf{\begin{tabular}[c]{@{}l@{}}RL TestBed\end{tabular}} & No            & No                    & Continuous       & No                 & No                \\ \hline
\textbf{Energym}                                                                 & Partial       & No                    & Continuous       & No                 & Yes               \\ \hline
\textbf{\begin{tabular}[c]{@{}l@{}}BOPTEST-Gym\end{tabular}}                  & No            & No                    & Both             & Yes                & Yes          \\ \hline
\textbf{\sinergym{}}                                                                & Yes           & Yes                   & Both             & Yes                & Yes               \\ 
\hline
\end{tabular}
\subcaption*{\textbf{WeatherSet}:  Can each building be used with different weather configurations? \textbf{WeatherVar}: Does the temperature dataset vary between episodes?
\textbf{Actions}: Which action spaces are supported? (discrete, continuous or both). \textbf{DynamicSp}: Dynamic spaces; can an environment be configured with different action and observation spaces than the predefined? \textbf{CustomRw}: Can custom rewards be defined?}
\end{table}

Note also how a \textit{Middleware} software is always required to convert continuous control actions into building actuator signals passed through the Gym/Gymnasium interface. RL Testbed patches the simulator source code, which makes it low scalable, difficult to update, and more difficult to use and adapt. Energym relies on the Functional Mock-up Interface (FMI)\footnote{\url{https://fmi-standard.org/literature/}}, which defines an interface to exchange dynamic simulation models aimed at co-simulation, similarly to the Gym interface. The BOPTEST project encompasses two components: a RESTful HTTP API based on the FMI that facilitates the simulation processes, and a Gymnasium-compliant interface to this API (namely, BOPTEST-Gym). In contrast, \sinergym{} uses the EnergyPlus Python API\footnote{\url{https://energyplus.readthedocs.io/en/latest/api.html}}, a recent extension of the simulator to access to its functionalities, making this interaction more efficient.

Regarding the supported buildings and configurations (\textit{Envs} column), every framework offers a list of pre-configured environments, each one including a building model and a weather specification, along with other features such as custom action and observation spaces. These models (\textit{Buildings}) define the architecture and equipment of the building, while the weather conditions are the specifications of the external temperatures and other related variables. Regarding these weather conditions (\textit{WeatherSet}), BOPTEST-Gym and RL Testbed do not allow their modification for a given building, whereas Energym and \sinergym{} allow the selection of different weather conditions for each building when setting up new environments. On top of this functionality, \sinergym{} implements mechanisms for weather variability (\textit{WeatherVar}), which are explained in detail in \autoref{subsec:weathertypes}.

RL Testbed and Energym only support continuous \textit{Actions}, whereas BOPTEST-Gym and \sinergym{} support continuous and discrete control. RL Testbed and Energym hardwire the allowed action values and different control specifications into the building definition files, which hinders the modification of actionable components and signals. In contrast, \sinergym{} and BOPTEST-Gym can programmatically modify several elements of the building definition (\textit{DynamicSp}) ---specifically, the observation and action spaces--- yielding different scenarios on the fly. 

A proper reward function is essential in the training process of RL agents, in order to assess the goodness of actions/states while avoiding undesirable side effects \cite{zhang2018b}. Since building models have different zones, sensors, actuators, etc., and different metrics to quantify successful control, the customization of reward functions is desirable (\textit{CustomRw}). RL Testbed's function reward is embedded into the code, thus hampering reward customization. BOPTEST supports the definition of key performance indicators (KPIs) via a RESTful service, which can be also used in the reward function of BOPTEST-Gym in addition to the predefined ones. Energym and \sinergym{} fully support customised rewards based on any observed variable, being directly specified when the environment is created.

CityLearn distinguishes itself from other frameworks due to its unique purpose and design. Like other tools, it addresses energy optimization, is implemented in Python, and supports RL through a Gymnasium interface. However, CityLearn is specifically aimed at benchmarking multi-agent RL controllers within a virtual smart grid composed of distributed energy resources. While the grid can include buildings, CityLearn abstracts their detailed characteristics and relies on a simplified demand response model, unlike the detailed simulation and control offered by other frameworks. Instead of using a physical simulation, CityLearn employs a data-driven predictive model to determine heating and cooling loads, utilizing a long short-term memory (LSTM) neural network \cite{pinto2021}. This eliminates the need for interaction with an external simulator, making CityLearn more self-contained. However, this design can make scenario development and reuse more challenging, particularly because CityLearn lacks a long-established community like those of Modelica and EnergyPlus, where extensive libraries, documentation, and support have accumulated over time. To address this, the authors provide a diverse set of preconfigured environments to facilitate use and experimentation.

To summarize, \sinergym{} stands out for its ability to integrate new simulated environments and extensively customise the simulation settings. BOPTEST-Gym demonstrates reliability and compatibility with Gymnasium, but it requires a deeper dive into lower-level abstractions for environments modifications, particularly regarding the observation and action spaces. Conversely, RL Testbed is more static, primarily designed for predefined testing scenarios. Finally, while Energym includes interesting features, lacks full compatibility with the Gym/Gymnasium API, thereby limiting its integration capabilities with external optimization and control libraries.
\section{Design}
\label{sec:design}

\subsection{Objectives}
\sinergym{} is designed to provide broad support for RL control, without excluding alternative control methods. The main objectives behind the design of the tool are the following:

\begin{itemize}
    
    \item \textbf{Flexibility to work with multiple and modifiable scenarios}. The software must support different buildings and weather configurations, allowing the selection and configuration of controllable elements of the building. Predefined scenarios should be provided for testing and benchmarking purposes.

    \item \textbf{Customizable data processing and model evaluation functions}. The tool must provide means to filter and read building variables to calculate key performance indicators. Furthermore, parametrized reward functions must be employed, together with structures for control actions and state representations allowing simple adaptation to different scenarios.

    \item \textbf{Large-scale experimentation and benchmarking support}. The tool must facilitate the execution of varied environments and guarantee the reproducibility of the results. Experiment parameters, metrics, and outputs must be tracked during the execution and recorded in an usable format for posterior analysis and visualization.

    \item \textbf{Adoption of high-quality standards of software development}. Scientific software sometimes lacks the ease of use and comprehensive documentation of professional tools, which is a barrier for both users and contributing developers. The software must be well-documented to engage a larger community, including illustrative examples that can be executed out of the box and adapted to specific use cases. Installation must be straightforward and based on modern dependency management tools.
\end{itemize}

We summarize these requirements in \autoref{table:requisites}, along with the corresponding functionalities of \sinergym{} that will be explained throughout \autoref{sec:functionalities}.

\begin{table}[h]
\caption{\sinergym{} design principles and functionalities}\label{table:requisites}
\renewcommand{\arraystretch}{1.2} 
\begin{adjustbox}{width=\textwidth}
\begin{tabular}{ll}
\toprule%
\textbf{Requirements} & \textbf{Functionalities} \\
\midrule
Flexibility to work with multiple and customizable environments & \begin{tabular}[c]{@{}l@{}}4.1. Building Models \\ 4.2. Weather Types\\ 4.3. Environments\end{tabular}\\ \hline
Customizable data processing and model evaluation functions & \begin{tabular}[c]{@{}l@{}}4.4. Rewards\\ 4.5. Wrappers\end{tabular}\\ \hline
Experimentation and benchmarking support & \begin{tabular}[c]{@{}l@{}}4.6. Controllers\\ 4.7. Benchmarking\end{tabular} \\ \hline
Adoption of high quality standards of software development & See repository \\
\hline
\end{tabular}
\end{adjustbox}
\end{table}

\subsection{Software architecture}
\label{subsec:design}
\begin{figure}
    \centering
    \includegraphics[width=\linewidth]{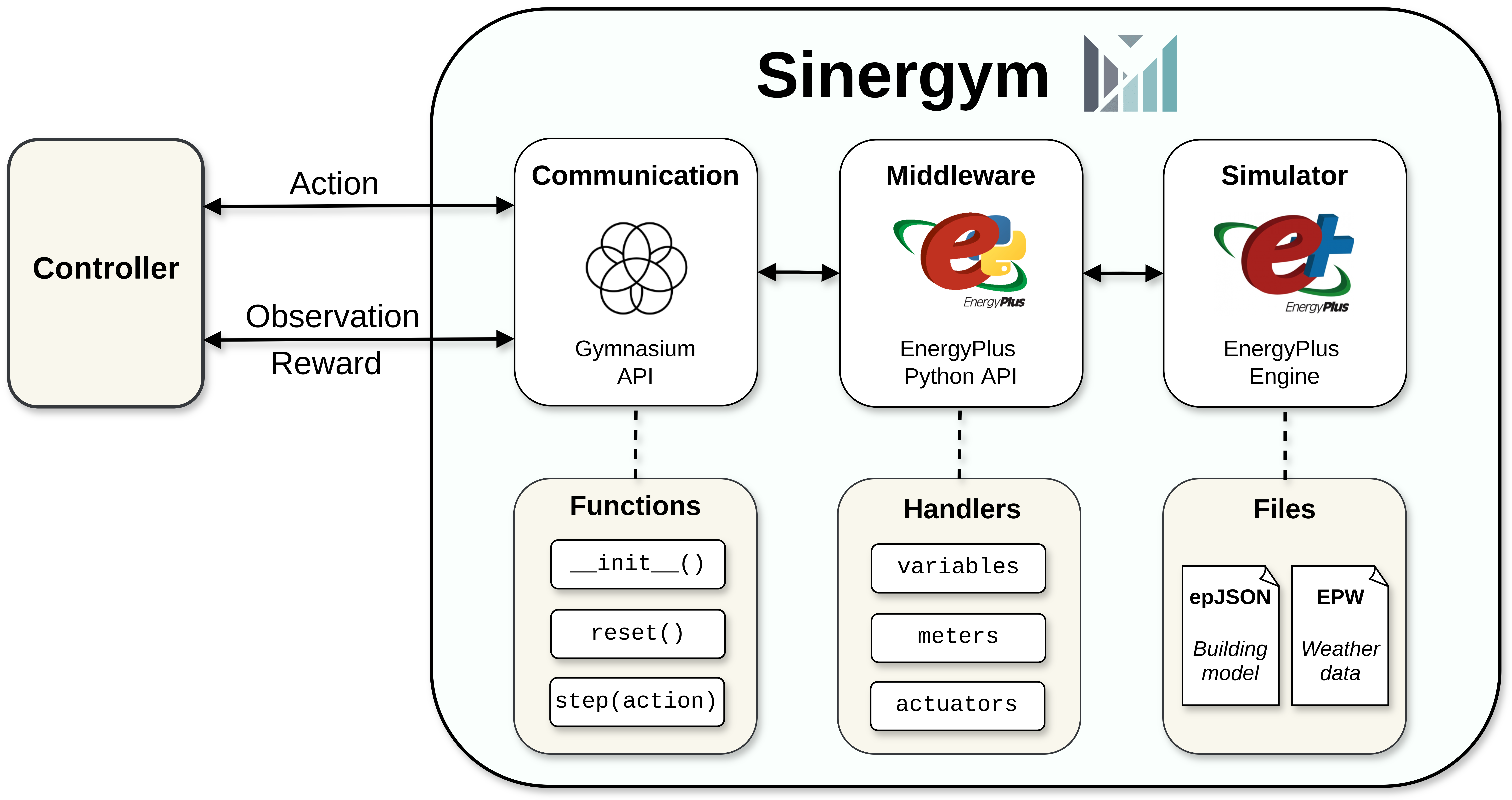}
    \caption{\sinergym{}'s general architecture, depicting its three main layers: \textit{communication}, \textit{middleware} and \textit{simulator}} 
    \label{fig:sinergym_architecture}
\end{figure}

The design of \sinergym{} is organized in three layers, as depicted in \autoref{fig:sinergym_architecture}: \textit{communication}, \textit{middleware}, and \textit{simulator}. The controller sends actions to and receives observations from the simulated building through the Gymnasium interface (\textit{communication} layer). The Gymnasium interface communicates with the simulation engine through the EnergyPlus Python API (\textit{middleware} layer). The EnergyPlus engine runs the simulation (\textit{simulator} layer) and updates the building state. The agent may use the information received from the environment to determine the subsequent control action and to modify its behaviour to maximize future rewards.

Every \sinergym{} simulation requires two data files to configure and launch the control process:

\begin{itemize}
    \item An \textbf{epJSON} (\textit{EnergyPlus JavaScript Object Notation}) file, which defines the building model simulated by EnergyPlus, as well as the controllable equipment components. Please refer to \autoref{subsec:buildingmodels} for further details.
    
    \item An \textbf{EPW} (\textit{EnergyPlus Weather}) file including the Typical Meteorological Year (TMY3) data to be used in the simulation. More details are provided in \autoref{subsec:weathertypes}.
\end{itemize}

\sinergym{} encapsulates the components implementing these three layers in an \texttt{Environment} class. This class implements the methods of the Gymnasium API, namely \texttt{init}, \texttt{reset} and \texttt{step}. Additional information about the interaction with \texttt{Environment} objects is provided in \autoref{subsec:environments}.

\begin{figure}
    \centering
    \includegraphics[width=\linewidth]{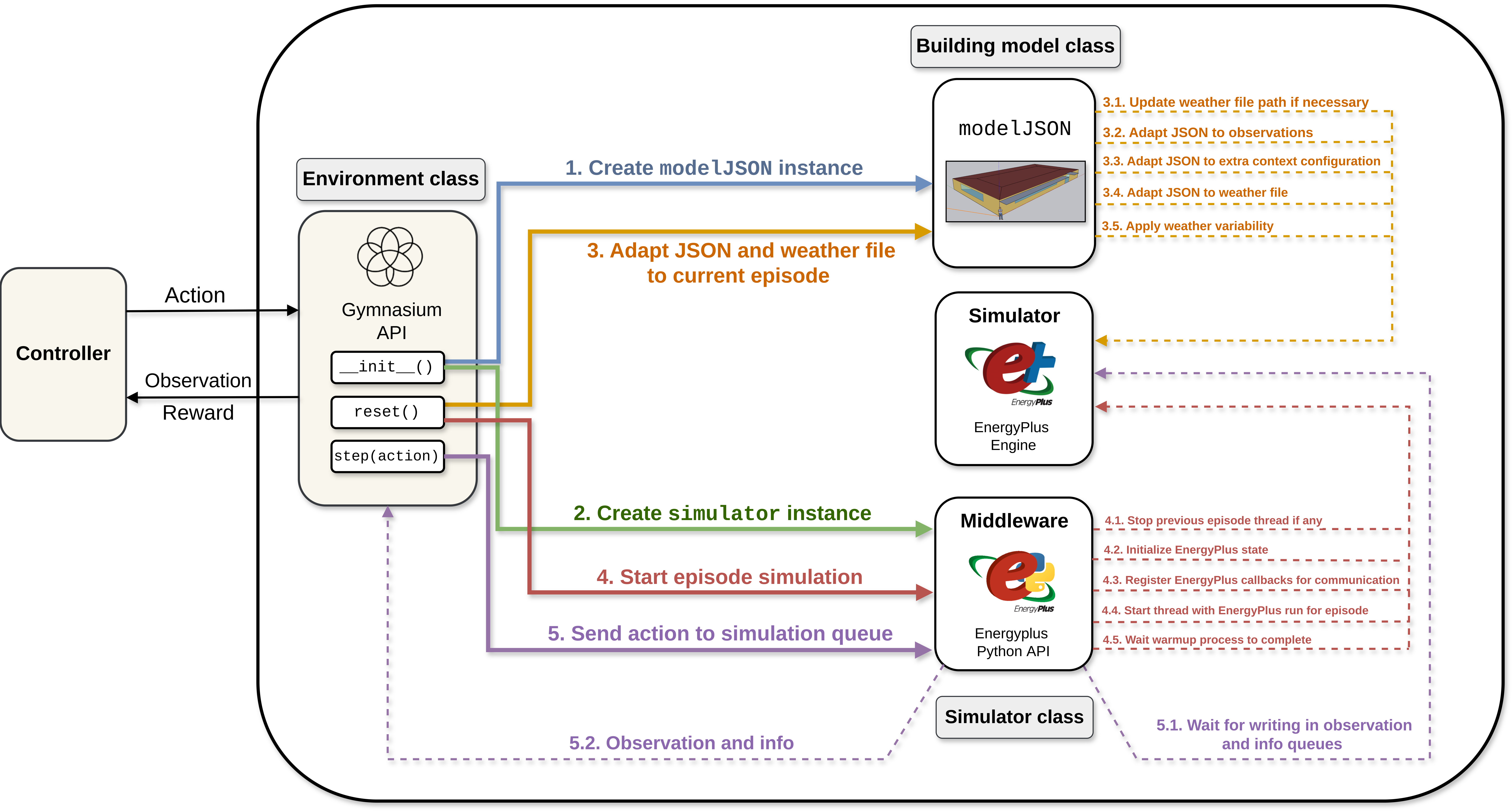}
    \caption{Overview of the \sinergym{} workflow}
    \label{fig:sinergym_backend}
\end{figure}

\autoref{fig:sinergym_backend} depicts in detail the internal process flow of the  \texttt{Environment} class of \sinergym{}. In step 1, objects are initialized with \texttt{\_\_init\_\_} using the epJSON file passed as a parameter. Next, in step 2, the simulator is launched through the EnergyPlus Python API. This step also involves the creation of handlers for the simulator, which are used to manage and communicate with the underlying engine. Now, the simulator is ready to be launched and the first episode can start after calling \texttt{reset}. Before starting each episode, \sinergym{} automatically configures some parameters of the simulation (step 3), e.g., noise application on the weather file, definition of building variables for sensors and actuators, or time configuration (steps per hour, run period, etc.). 

Step 4 comprises the stages of the simulation cycle: stopping the previous run, if any; setting up the initial state; warm-up process; and the simulation along the run period. While the simulation is running, a controller can interact with the building (step 5). \sinergym{} overrides the default building control by overwriting the default EnergyPlus schedulers. When actions are sent to the simulation by calling \texttt{step}, \sinergym{} transparently interrupts the simulation, sets the control signal, and resumes the process.

\section{Functionalities}
\label{sec:functionalities}
This section provides an overview of the key functionalities of \sinergym{} 3.6.2. For updated information and examples, please refer to its official documentation website\footnote{\url{https://ugr-sail.github.io/sinergym}}.

\subsection{Building models}\label{subsec:buildingmodels}
Building models listed in \autoref{table:buildings} are included by default in \sinergym{} and are ready to use. Additional experimental buildings and comprehensive information can be found in the project documentation website. The underlying epJSON files are reference models provided by the American Society of Heating, Refrigerating and Air-Conditioning Engineers (ASHRAE)\footnote{\url{https://www.ashrae.org/}} or supported by the U.S. Department of Energy (DOE)\footnote{\url{https://www.energycodes.gov/prototype-building-models}}. 

From these buildings, \sinergym{} allows the creation of variants on the fly (see \autoref{subsec:environments}), thus avoiding the manual duplication and edition of epJSON files. These variants include changes to a subset of the model components, which are programmatically defined when the environment is created and internally materialized. Furthermore, \sinergym{} offers additional functionalities to facilitate the integration of new building models, such as the automatic detection of sensors and actuators in epJSON files, and the subsequent preparation of these files to be accessed through the Gymnasium interface.

\begin{table}[htpb]
\caption{Description of the base building models included with \sinergym{}}\label{table:buildings}
\centering
\begin{adjustbox}{width=0.95\linewidth}
\begin{tabular*}{\textwidth}{@{\extracolsep\fill}p{0.7\textwidth}  p{0.3\textwidth}}
\toprule%
\textbf{Building} & \textbf{Appearance}\\
\midrule
\textbf{5ZoneAutoDXVAV} \cite{ding2020,wei2017}: A single-story building divided into 1 indoor and 4 outdoor zones. There are windows on all 4 facades. Its floor area is 463.6 m$^2$, and it is equipped with a packaged Variable Air Volume (VAV) system, Direct Expansion (DX) cooling coil, and gas heating coils, with fully auto-sized input. & \begin{center}\includegraphics[width=0.3\textwidth]{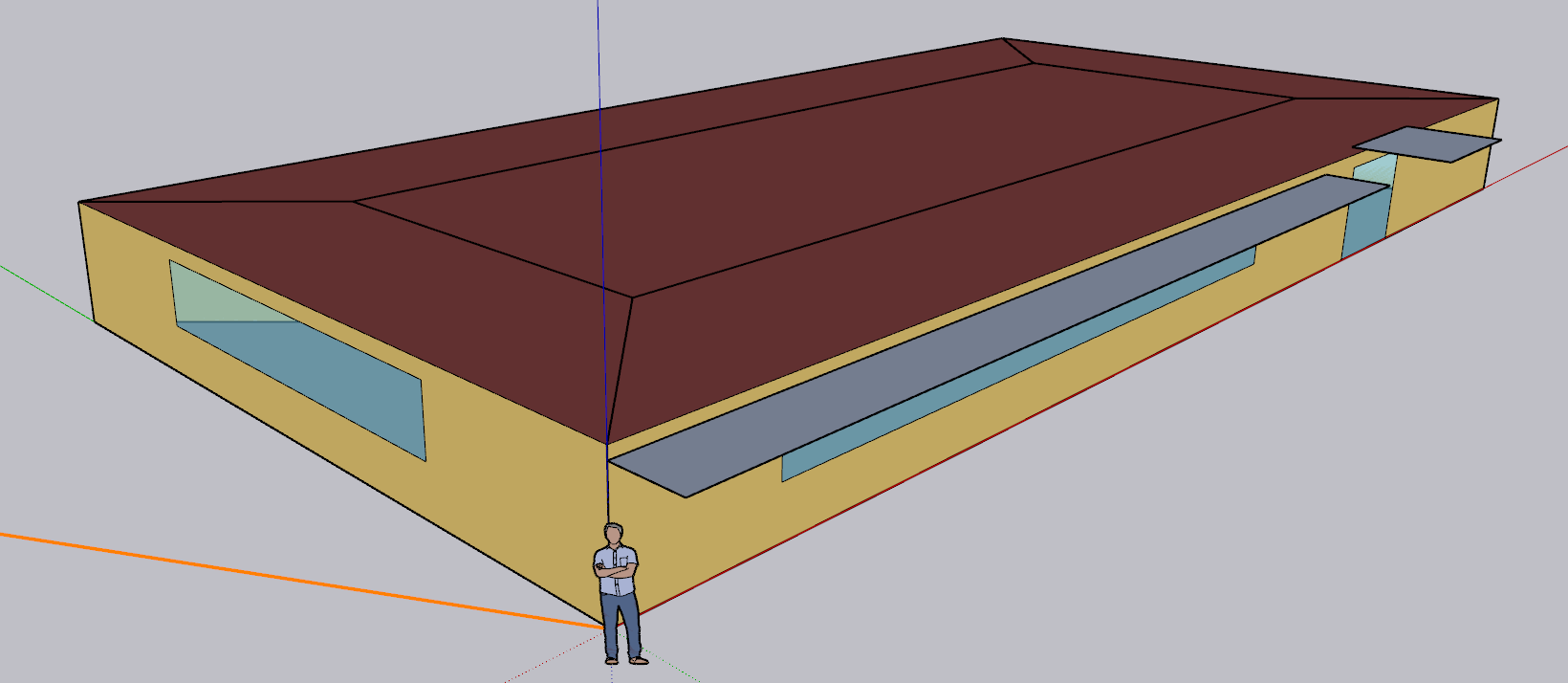}\end{center}\\

\textbf{2ZoneDataCenterHVAC\_wEconomizer} \cite{li2020,moriyama2018}: A 491.3 m$^2$ building divided into two asymmetrical zones (west and east zone). Each zone has an HVAC system consisting of air economizers, evaporative coolers, DX cooling coil, chilled water coil, and VAV units. The main heat source comes from the hosted servers. The heating and cooling setpoint values are applied simultaneously to both zones. & \begin{center}\includegraphics[width=0.3\textwidth]{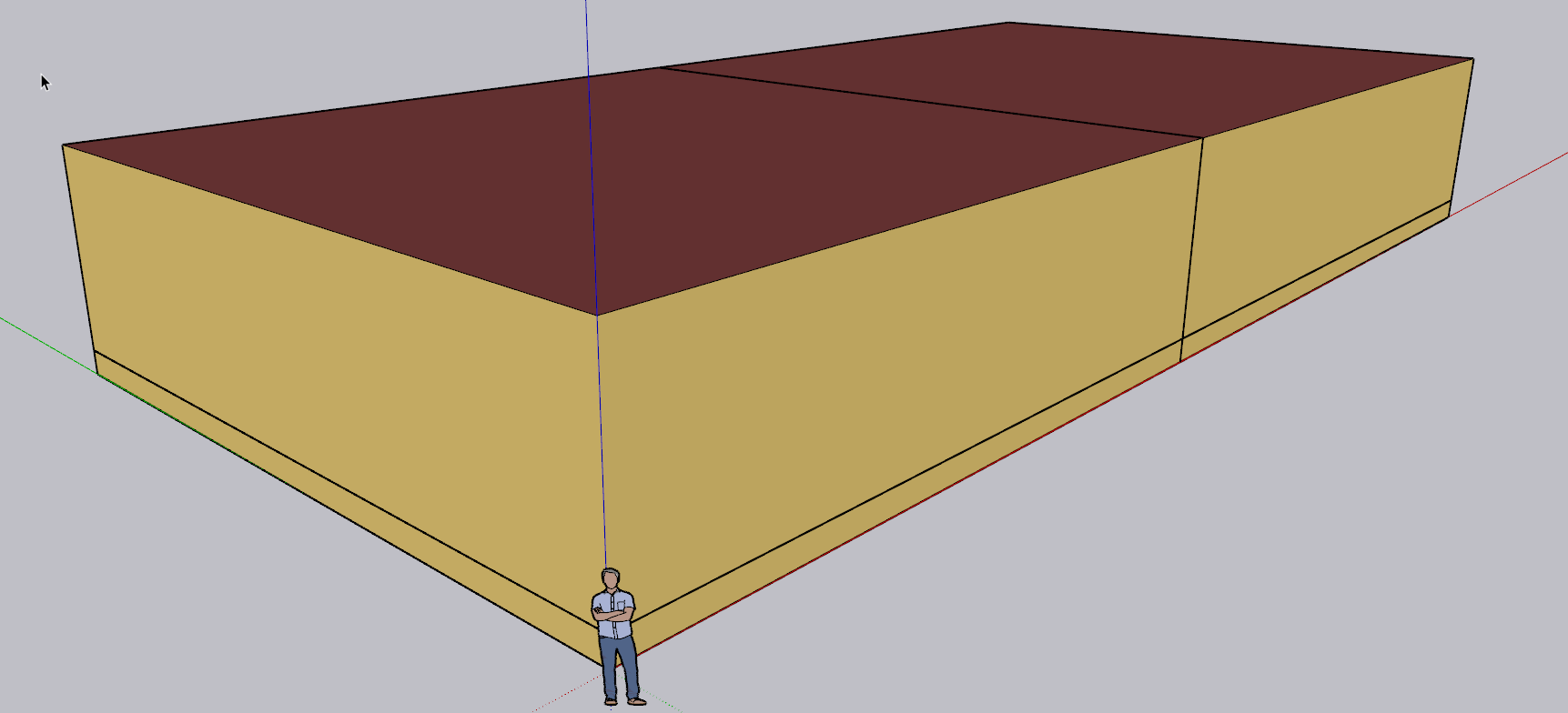} \end{center}\\

\textbf{ASHRAE9012016\_OfficeMedium} \cite{pinto2021,cho2010}: A 4,979.6 m2 building with three floors and  equipped with an HVAC system. Each floor has four perimetral zones and one core zone. Available fuel types are gas and electricity. & \begin{center}\includegraphics[width=0.3\textwidth]{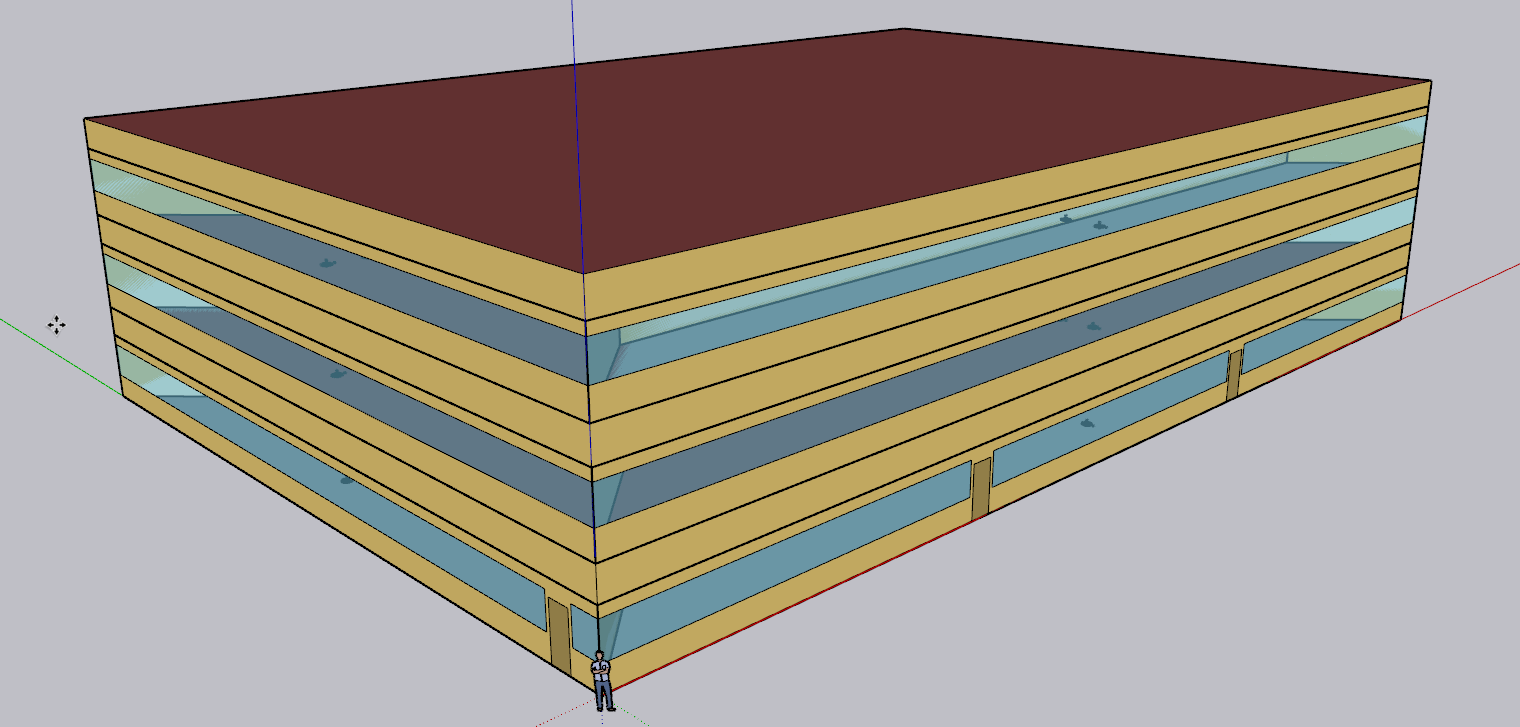} \end{center}\\ 

\textbf{ASHRAE9012016\_Warehouse} \cite{cho2010}: A non-residential single-storey building with 4,598 m$^2$, equipped with an HVAC system. The building is divided into three zones: bulk storage, fine storage, and office. The office area is enclosed by the fine storage zone, and it is the only zone with windows. The available fuel types are gas and electricity. & \begin{center}\includegraphics[width=0.3\textwidth]{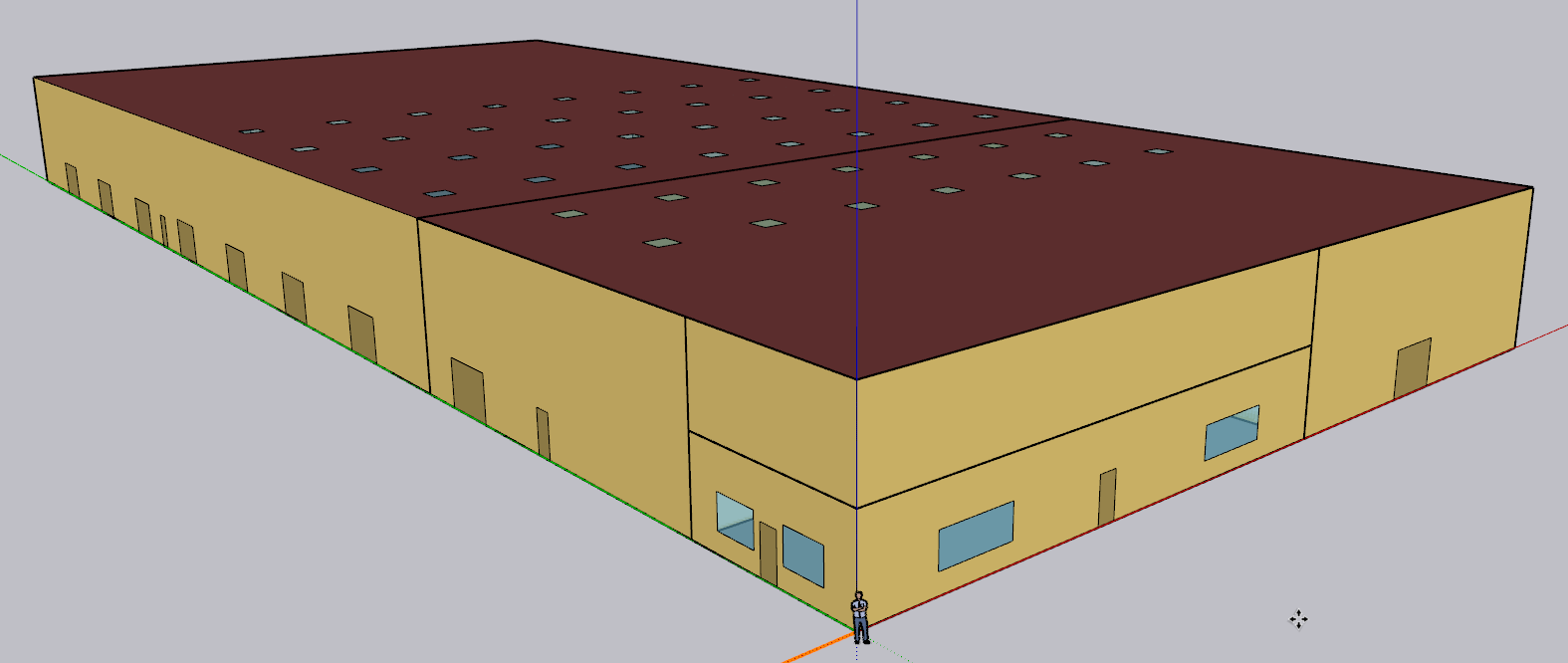} \end{center} \\ 
\hline
\end{tabular*}
\end{adjustbox}
\end{table}

\subsection{Weather types}
\label{subsec:weathertypes}
In EnergyPlus, weather data is decoupled from the building models and are read from independent EPW files. \sinergym{} includes different weather files provided by EnergyPlus, covering varied climate regions (see \autoref{table:weathers}). It also includes functionalities to modify weather data during simulation time, and to integrate custom weather files. 

\begin{table}[H]
\caption{Weather types available in \sinergym{}. MAT: mean air temperature; MAH: mean air humidity}\label{table:weathers}
\centering
\renewcommand{\arraystretch}{1}
\begin{adjustbox}{width=0.8\textwidth}
\begin{tabular}{lllll}
\toprule
\textbf{Weather file}                                                                            & \textbf{Location}                                                   & \textbf{Climate type}                                                                                                                      & \textbf{MAT (\textcelsius)} & \textbf{MAH (\%)} \\ 
\midrule
\textbf{AUS\_NSW.Sydney}                                                                         & \begin{tabular}[c]{@{}l@{}}Sydney, \\ Australia\end{tabular}        & \begin{tabular}[c]{@{}l@{}}Humid subtropical\\ (no dry seasons and \\ hot summers).\end{tabular}                                           & 17.9              & 68.83             \\ \hline
\textbf{COL\_Bogota}                                                                             & \begin{tabular}[c]{@{}l@{}}Bogota, \\ Colombia\end{tabular}         & \begin{tabular}[c]{@{}l@{}}Mediterranean \\ (dry, warm \\ summers and mild \\ winters).\end{tabular}                                       & 13.2              & 80.3              \\ \hline
\textbf{ESP\_Granada}                                                                            & \begin{tabular}[c]{@{}l@{}}Granada, \\ Spain\end{tabular}           & \begin{tabular}[c]{@{}l@{}}Mid-latitude dry semiarid \\ and hot dry periods \\ in summer, but passive \\ cooling is possible.\end{tabular} & 14.84             & 59.83             \\ \hline
\textbf{FIN\_Helsinki}                                                                           & \begin{tabular}[c]{@{}l@{}}Helsinki, \\ Finland\end{tabular}        & \begin{tabular}[c]{@{}l@{}}Moist continental (warm \\ summers, cold winters, \\ no dry seasons).\end{tabular}                              & 5.1               & 79.25             \\ \hline
\textbf{\begin{tabular}[c]{@{}l@{}}JPN\_\\ Tokyo.Hyakuri\end{tabular}}                           & \begin{tabular}[c]{@{}l@{}}Tokyo, \\ Japan\end{tabular}             & \begin{tabular}[c]{@{}l@{}}Humid subtropical (mild \\ with no dry season, \\ hot summer).\end{tabular}                                     & 8.9               & 78.6              \\ \hline
\textbf{\begin{tabular}[c]{@{}l@{}}MDG\_\\ Antananarivo\end{tabular}}                            & \begin{tabular}[c]{@{}l@{}}Antananarivo, \\ Madagascar\end{tabular} & \begin{tabular}[c]{@{}l@{}}Mediterranean climate \\ (dry warm summer, \\ mild winter).\end{tabular}                                        & 18.35             & 75.91             \\ \hline
\textbf{PRT\_Lisboa}                                                                             & \begin{tabular}[c]{@{}l@{}}Lisboa, \\ Portugal\end{tabular}         & \begin{tabular}[c]{@{}l@{}}Dry Summer Subtropical \\ Mediterranean (Warm - \\ Marine).\end{tabular}                                        & 16.3              & 74.2              \\ \hline
\textbf{\begin{tabular}[c]{@{}l@{}}USA\_AZ\_Davis-\\ Monthan\end{tabular}}                       & \begin{tabular}[c]{@{}l@{}}Arizona, \\ USA\end{tabular}             & \begin{tabular}[c]{@{}l@{}}Subtropical hot desert \\ (hot and dry).\end{tabular}                                                           & 21.7              & 34.9              \\ \hline
\textbf{\begin{tabular}[c]{@{}l@{}}USA\_CO\_Aurora-\\ Buckley\end{tabular}}                      & \begin{tabular}[c]{@{}l@{}}Colorado, \\ USA\end{tabular}            & \begin{tabular}[c]{@{}l@{}}Mid-latitude dry semiarid \\ (cool and dry).\end{tabular}                                                       & 9.95              & 55.25             \\ \hline
\textbf{\begin{tabular}[c]{@{}l@{}}USA\_IL\_Chicago-\\ OHare\end{tabular}}                       & \begin{tabular}[c]{@{}l@{}}Illinois, \\ USA\end{tabular}            & \begin{tabular}[c]{@{}l@{}}Humid continental \\ (mixed and humid).\end{tabular}                                                            & 9.92              & 70.3              \\ \hline
\textbf{\begin{tabular}[c]{@{}l@{}}USA\_NY\_\\ New.York-\\ J.F.Kennedy\end{tabular}}             & \begin{tabular}[c]{@{}l@{}}New York, \\ USA\end{tabular}            & \begin{tabular}[c]{@{}l@{}}Humid continental \\ (mixed and humid).\end{tabular}                                                            & 12.6              & 68.5              \\ \hline
\textbf{\begin{tabular}[c]{@{}l@{}}USA\_PA\_\\ Pittsburgh\end{tabular}}                          & \begin{tabular}[c]{@{}l@{}}Pennsylvania, \\ USA\end{tabular}        & \begin{tabular}[c]{@{}l@{}}Humid continental \\ (cool and humid).\end{tabular}                                                             & 10.5              & 66.41             \\ \hline
\textbf{\begin{tabular}[c]{@{}l@{}}USA\_WA\_\\ Port.Angeles-\\ William.R.Fairchild\end{tabular}} & \begin{tabular}[c]{@{}l@{}}Washington, \\ USA\end{tabular}          & \begin{tabular}[c]{@{}l@{}}Cool Marine west \\ coastal (warm summer, \\ mild winter, rain all year).\end{tabular}                          & 9.3               & 81.1              \\ 
\hline
\end{tabular}
\end{adjustbox}
\end{table}

\begin{figure}
    \centering
    \includegraphics[width=\linewidth]{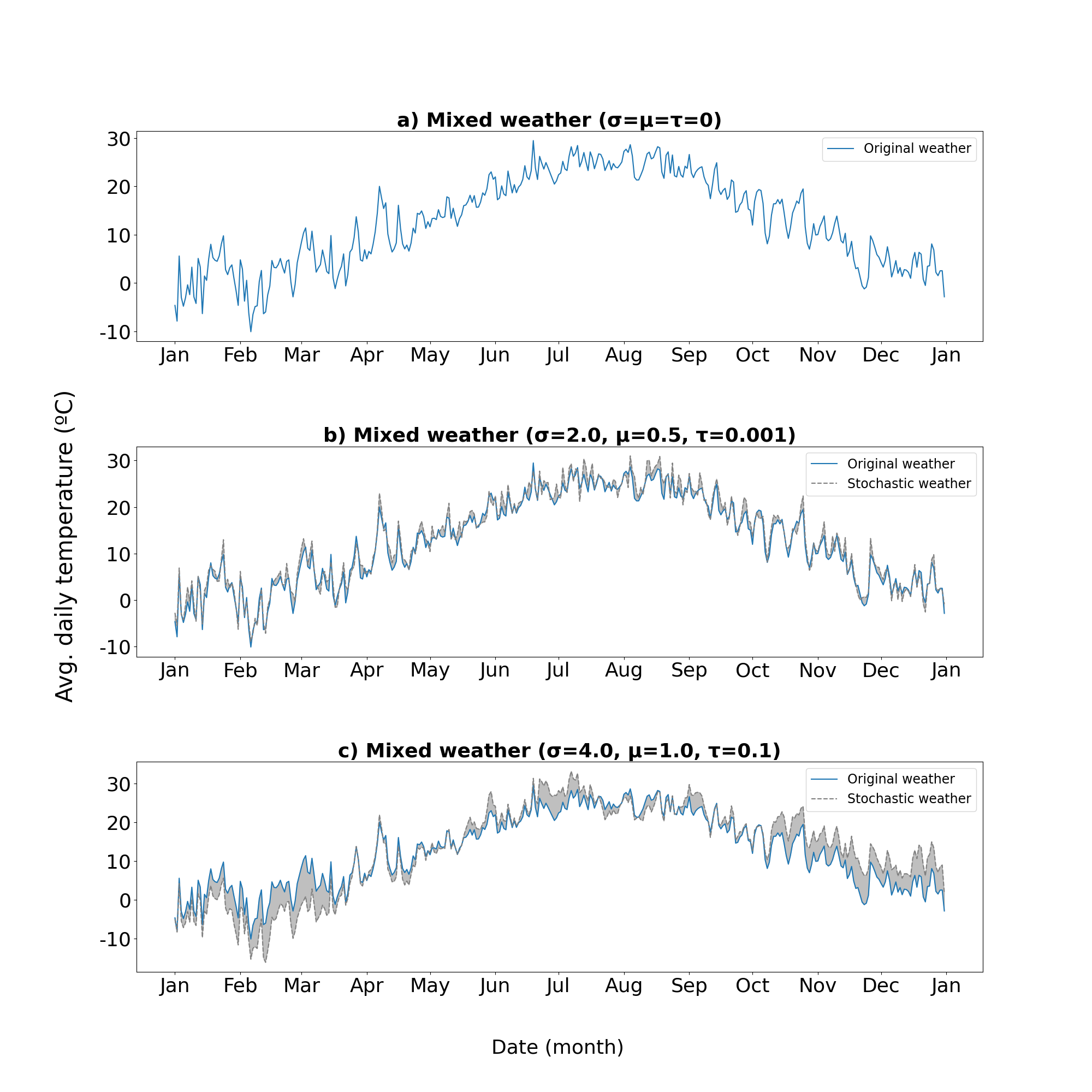}
    \caption{Ornstein-Uhlenbeck noise applied to a mixed weather dataset (New York) with different values for $\sigma$, $\mu$ and $\tau$}
    \label{fig:weather_variability}
\end{figure}

As shown in \autoref{fig:weather_variability}, \sinergym{} implements weather variations by adding noise to the recorded TMY3 temperature. This noise is applied by using an Ornstein-Uhlenbeck (OU) process \cite{zarate2013}, which is similar to a random walk in continuous time. Following \autoref{eq:equation_weather}, the temperature at the next timestep $T_{t+1}$ is calculated from the previous $T_t$, with parameters $\sigma$ (standard deviation of the signal), $\mu$ (mean of the signal), $\tau$ (time constant), and $W(t_{i+1}) - W(t_i) \sim \mathcal{N}(0, t_{i+1} - t_{i})$ (Gaussian noise).

\begin{equation}
\label{eq:equation_weather}
    T_{t+1}=(1-\mu)T_t-\tau+\sigma(W_{t+1}-W_t)
\end{equation}

\subsection{Scenarios and environments}
\label{subsec:environments}

\sinergym{} provides 87 scenarios, each one corresponding to a combination of an epJSON file, an EPW file, selected action and observation spaces, and a predefined reward function. The scenarios can be loaded as environments by specifying the parameter values listed in \autoref{table:parameters}. The software automatically creates all the necessary supporting files and data structures, such as the adapted epJSON building file, the weather file with perturbations (if enabled), and the available handlers for the actionable elements of the building. 

\begin{table}[H]
\centering
\caption{Configurable parameters in the creation of a \sinergym{} environment}
\label{table:parameters}
\begin{adjustbox}{width=\textwidth}
\begin{tabular}{llll}
\toprule
\textbf{\begin{tabular}[c]{@{}l@{}}Parameter\end{tabular}}              & \textbf{\begin{tabular}[c]{@{}l@{}}Type\end{tabular}}            & \textbf{Optional?} & \textbf{Description}
\\ \midrule
\textit{building\_file}                                                        & \texttt{str}                                                                      & No                & Name of the \texttt{epJSON} file with the default building definition.                                                                                                                                                                           \\ \hline
\textit{weather\_files}                                                        & \begin{tabular}[c]{@{}l@{}}\texttt{str} \\ or\\ \texttt{List{[}str{]}}\end{tabular}        & No                & \begin{tabular}[c]{@{}l@{}}Name of the \texttt{EPW} file with weather data. A list of \\ weather files can be specified from which to sample \\ at each episode.\end{tabular}                                     \\ \hline
\textit{action\_space}                                                         & \begin{tabular}[c]{@{}l@{}}\texttt{gym.spaces.}-\\ \texttt{Box} or\\ \texttt{Discrete}\end{tabular} & Yes               & \begin{tabular}[c]{@{}l@{}}Gymnasium action space definition. Defaults to an empty action \\ space, which corresponds to the baseline building controller.\end{tabular}                                                                   \\ \hline
\textit{time\_variables}                                                       & \texttt{List{[}str{]}}                                                            & Yes               & \begin{tabular}[c]{@{}l@{}}EnergyPlus time variables to be observed. The variable \\ name must match the name of the EnergyPlus Data Transfer \\ API method. Defaults to an empty list.\end{tabular}               \\ \hline
\textit{variables}                                                             & \texttt{Dict}                                                                     & Yes               & \begin{tabular}[c]{@{}l@{}}Specification for EnergyPlus \texttt{Output:Variable}. \\ The key name is user-defined, and the tuple must be the \\ original variable name and the output variable key. \\ Defaults to an empty dictionary.\end{tabular}       \\ \hline
\textit{meters}                                                                & \texttt{Dict}                                                                       & Yes               & \begin{tabular}[c]{@{}l@{}}Specification for EnergyPlus \texttt{Output:Meter}. The key \\ name is user-defined, and the value is the original EnergyPlus \\ Meters name. Defaults to an empty dictionary.\end{tabular}                                     \\ \hline
\textit{actuators}                                                                      & \texttt{Dict}                                                                       & Yes               & \begin{tabular}[c]{@{}l@{}}Specification for EnergyPlus actuators. The \\ key name is user-defined, and the value is a tuple with \\ actuator type, value type and original actuator name.\\ Defaults to an empty dictionary.\end{tabular} \\ \hline                                                                                                        
\textit{\begin{tabular}[c]{@{}l@{}}weather\_-\\ variability\end{tabular}}      & \texttt{Tuple{[}float{]}}                                                         & Yes               & \begin{tabular}[c]{@{}l@{}}Tuple with $\sigma$, $\mu$ and $\tau$ of the Ornstein-Uhlenbeck process \\ to be applied to weather data. Defaults to \texttt{None}.\end{tabular}                                                             \\ \hline
\textit{reward}                                                                & \texttt{Reward Class}                                                             & Yes               & \begin{tabular}[c]{@{}l@{}}Reward function class. \\ Defaults to \texttt{LinearReward}.\end{tabular}                                                                                                             \\ \hline
\textit{reward\_kwargs}                                                        & \texttt{Dict}                                                                     & Yes               & \begin{tabular}[c]{@{}l@{}}Parameters to be passed to the reward function. \\ Varies depending on the reward class. Defaults to an empty dictionary.\end{tabular}                                                                  \\ \hline
\textit{\begin{tabular}[c]{@{}l@{}}max\_ep\_data\_-\\ store\_num\end{tabular}} & \texttt{int}                                                                      & Yes               & \begin{tabular}[c]{@{}l@{}}Number of last sub-folders with episodic data \\ saved during simulation. Defaults to 10.\end{tabular}                                                                   \\ \hline
\textit{env\_name}                                                             & \texttt{str}                                                                      & Yes               & \begin{tabular}[c]{@{}l@{}}Environment name used for working directory creation. \\ Defaults to \texttt{`eplus-env-v1'}.\end{tabular}                                                                                                          \\ \hline
\textit{config\_params}                                                        & \texttt{Dict}                                                                     & Yes               & \begin{tabular}[c]{@{}l@{}}Dictionary with extra configuration for the simulator, \\ e.g., maximum simulation time, or environment sampling frequency.\\ Defaults to \texttt{None}.\end{tabular} \\ \hline                                      
\end{tabular}
\end{adjustbox}
\end{table}

Two key elements of the environment are: the observation space, which maps variables to the building sensors, and the action space, which maps variables to the building actuators. In particular, the observation space incorporates meters, dates and other output variables from EnergyPlus, while the action space aligns with the EnergyPlus input actuators. Notice that when an actuator is not defined when the \texttt{Environment} object is created, \sinergym{} uses the default schedulers defined in the base epJSON file of the building model.

\subsection{Reward functions}
\label{subsec:rewards}
 \sinergym{} allows users to select between several predefined reward functions and to implement their own custom rewards. If not specified, the default reward function linearly combines energy consumption and thermal discomfort with user-defined weights and comfort ranges, as represented by \autoref{eq:linear-reward}:

\begin{equation}
\label{eq:linear-reward}
     r_t = - \omega \ \lambda_P \ P_t - (1 - \omega) \ \lambda_T \ (|T_t - T_{up}| + |T_t - T_{low}|)
\end{equation}

where $P_t$ represents power consumption (W); $T_t$ is the current indoor temperature (\textcelsius); $T_{up}$ and $T_{low}$ are the imposed comfort range limits (penalty is $0$ if $T_t$ is within this range); $\omega$ is the weight assigned to power consumption (and consequently, $1 - \omega$, the comfort weight), and $\lambda_P$ and $\lambda_T$ are scaling constants for consumption and comfort, respectively. Note how the reward function is expressed in negative terms, as we look for its maximization.

\sinergym{} includes different alternatives to this default linear reward. For instance, in the \textit{exponential reward}, the discomfort penalty is calculated by an exponential function, thus increasing the cost of out-of-range indoor temperatures. The \textit{schedule reward} uses an additional weighting to modulate the discomfort term, which varies depending on the building occupancy. Furthermore, users can create more sophisticated reward functions involving building-specific variables during the environment definition step. 

\begin{figure}[h]
    \centering
    \includegraphics[width=\linewidth]{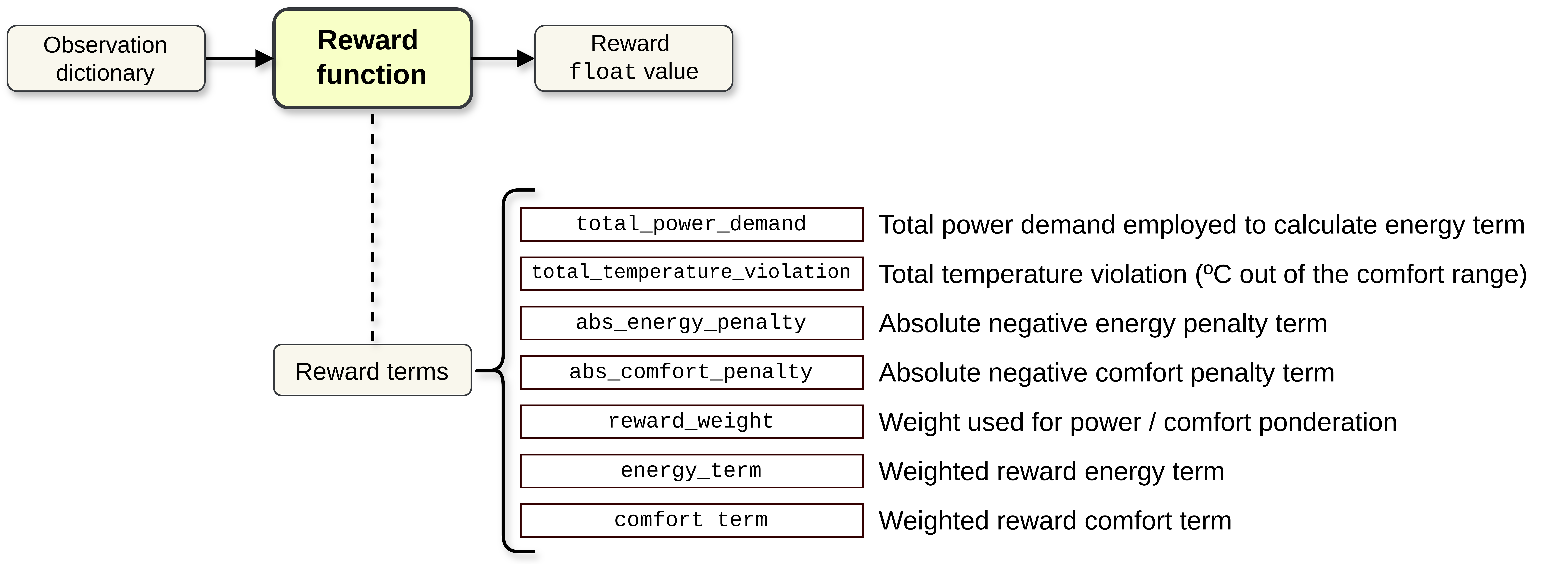}
    \caption{Example of intermediate values computed by the reward function}
    \label{fig:reward_terms}
\end{figure}

Finally, as illustrated in \autoref{fig:reward_terms}, the implementation of the reward function also makes the values of the terms used in its calculation available.

\subsection{Wrappers}
\label{subsec:wrappers}
Gymnasium wrappers are a convenient way to enrich the interaction with an environment without modifying it. More precisely, wrappers are used to process the data exchanged between the agent and the environment without coding. \sinergym{} includes many commonly used wrappers, such as observation normalization, action discretization, incremental control, observation stacking and logging. These wrappers can be nested to create complex processing pipelines that favour the training and monitoring of the control agent.

\subsection{Controllers}
\label{subsec:controllers}
Every controller in \sinergym{} is a process that sends actions to the environment through the \texttt{step} function. Therefore, it can be a random controller, a reactive rule-based system, an RL agent, etc. By default, \sinergym{} adapts to any controller aligned with the Gymnasium interface. This includes controllers defined in the epJSON files, manually-defined controllers, or even complex DRL implementations. Adding custom or learnt controllers  ---particularly by using libraries compliant with the Gymnasium API--- is simple and straightforward.

\subsection{Benchmarking}
\label{subsec:drl}
\sinergym{} supports logging and enables the comparison between multiple experiments' outputs. These favours reproducible results and promote fair comparisons between controllers. Specifically, environments can be monitored either during the training or evaluation of the controllers, by enabling pre-implemented function callbacks. By using platforms such as Weights\&Biases \cite{biewald2020}, users can define the hyperparameters, loggers, and data storage configuration of the trained models, as well as perform visual analysis through a web-based dashboard. The documentation of \sinergym{} provides templates to guide the definition and execution of RL training, and to evaluate experiments with Weights\&Biases.

\subsection{Installation and running}
The installation and configuration of the software are effortless by using PyPi\footnote{\url{https://pypi.org/project/sinergym/}} or the Docker distribution\footnote{\url{https://hub.docker.com/r/sailugr/sinergym}}. Consequently, \sinergym{} can run on a local machine or a cloud computing infrastructure.

\section{Examples}
\label{sec:examples}
This section showcases the practical application of \sinergym{} 3.6.2 through three BEO use cases: 1) testing the default control of a EnergyPlus building model, 2) using a custom rule-based controller, 3) training DRL controllers from scratch, 4) hyperparameter optimization of the DRL algorithm. The environment is the predefined \texttt{Eplus-datacenter-mixed-continuous-stochastic-v1}, featuring the \datacenter{} building, mixed weather (\textit{USA\_NY\_New.York-J.F.Kennedy} from \autoref{table:weathers}) with noise-added outdoor temperature, and continuous actions. The description of the action and observation spaces, wrappers, reward function, and hyperparameters are included in \ref{appendix:examples_details}.

Although the main objective of these examples is not to find the best controller for the given control problem, but to show the possibilities of the tool, we show how DRL agents were able to identify effective control strategies without the need for expert knowledge or detailed modelling.

\begin{figure}[p]
    \makebox[\textwidth][c]{%
    \begin{minipage}{1.2\textwidth}  
    \centering
    \begin{subfigure}[b]{0.48\textwidth}
        \centering
        \includegraphics[width=\textwidth]{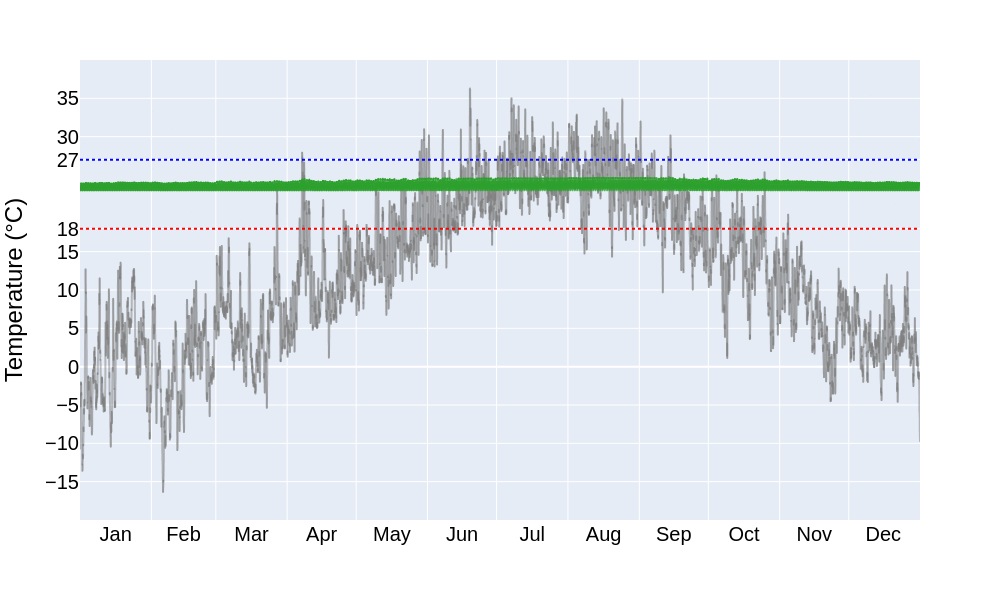}
        \caption{Default controller}
        \label{fig:temperatures_default}
    \end{subfigure}
    \hfill  
    \begin{subfigure}[b]{0.48\textwidth}
        \centering
        \includegraphics[width=\textwidth]{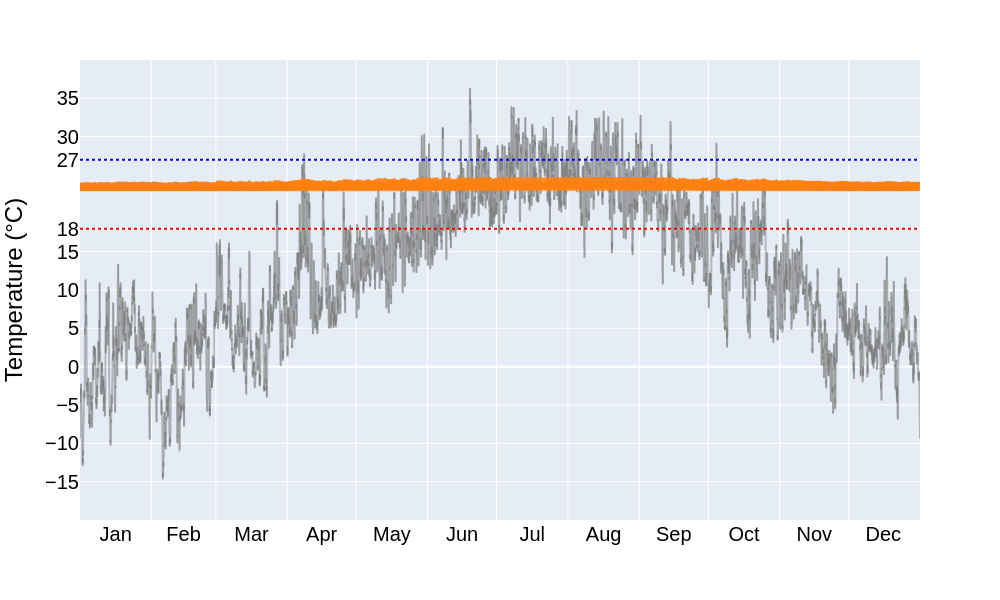}
        \caption{RBC}
        \label{fig:temperatures_rbc}
    \end{subfigure}
    
    \vspace{0.5cm}  

    \begin{subfigure}[b]{0.48\textwidth}
        \centering
        \includegraphics[width=\textwidth]{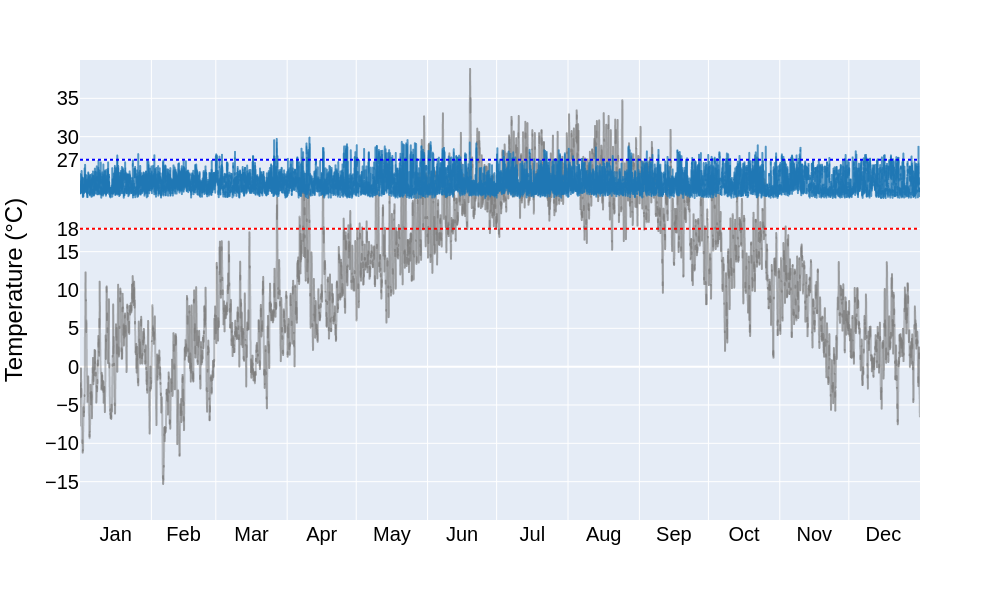}
        \caption{SAC}
        \label{fig:temperatures_sac}
    \end{subfigure}   
    \hfill
    \begin{subfigure}[b]{0.48\textwidth}
        \centering
        \includegraphics[width=\textwidth]{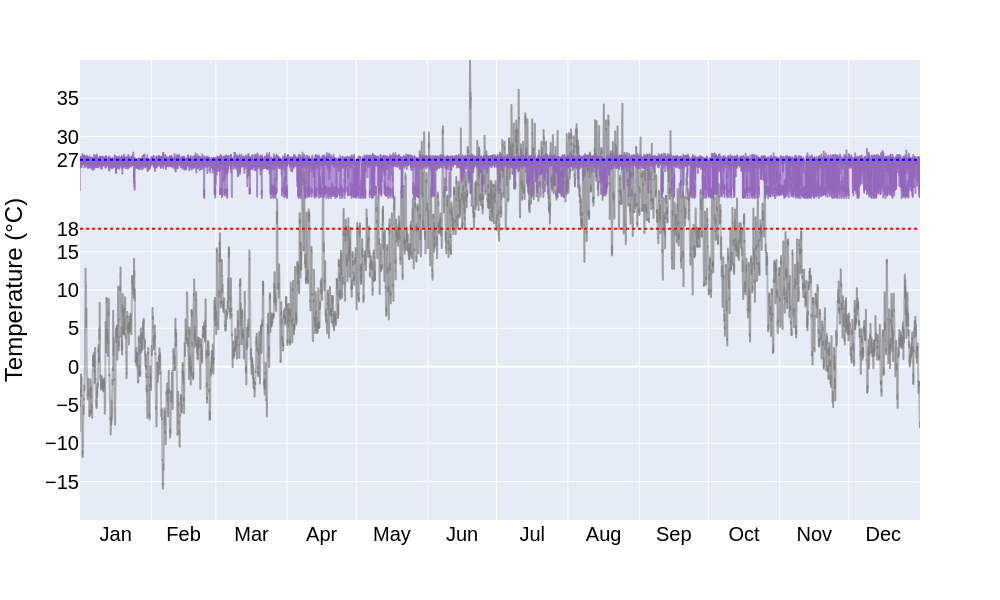}
        \caption{TD3}
        \label{fig:temperatures_td3}
    \end{subfigure}
    
    \vspace{0.5cm}

    \begin{subfigure}[b]{0.48\textwidth}
        \centering
        \includegraphics[width=\textwidth]{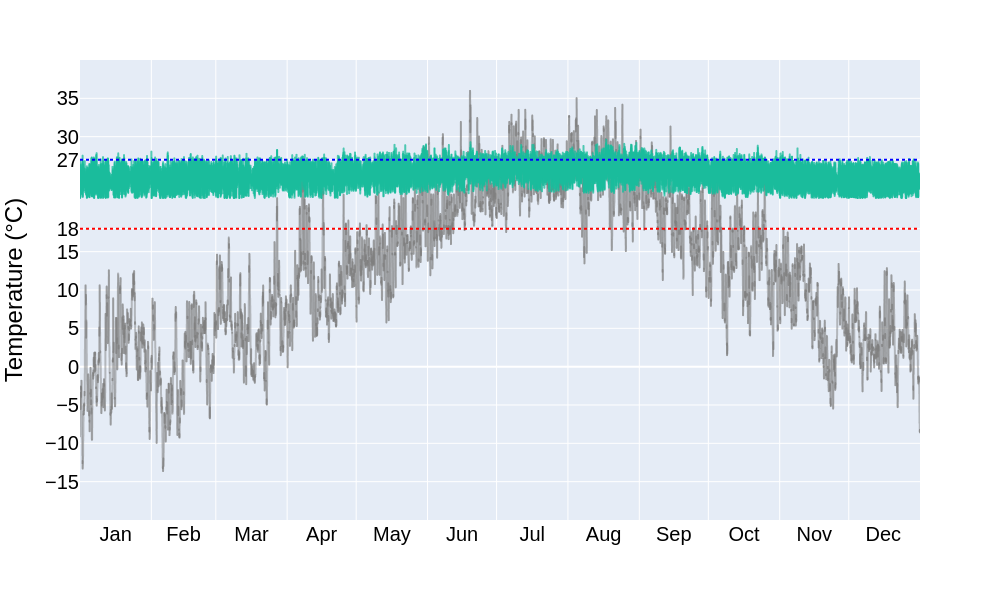}
        \caption{Baseline PPO}
        \label{fig:temperatures_ppo}
    \end{subfigure}
    \hfill
    \begin{subfigure}[b]{0.48\textwidth}
        \centering
        \includegraphics[width=\textwidth]{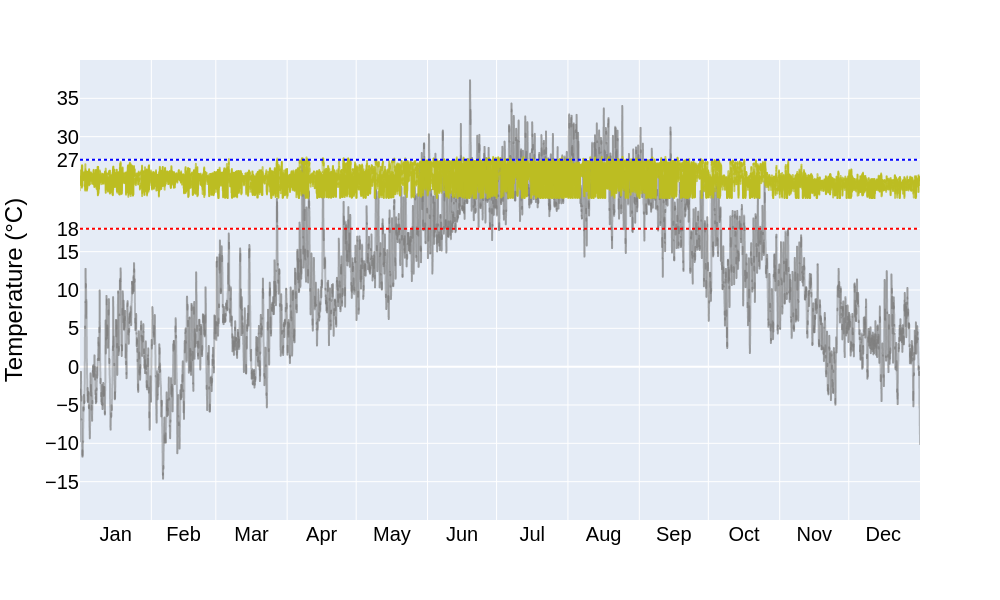}
        \caption{Tuned PPO}
        \label{fig:temperatures_ppo_tuned}
    \end{subfigure}
    
    \caption{Indoor air temperatures achieved by each controller for a 1-year evaluation period in \datacenter{} with mixed weather. The red and blue dotted straight lines indicate the comfort range. Indoor air temperatures are shown in colour, while outdoor temperatures are shown in grey.}
    \label{fig:all_experiments_temp}
    \end{minipage}%
    }
\end{figure}

\subsection{Example~1: Default building controller}
\label{subsec:example1}
In this case, the environment is instantiated with an empty action space, in order to use the default control defined in the building model file.  The default control for the \datacenter{} temperature setpoints is straightforward, as it statically sets the heating setpoint to 20 \textcelsius, and the cooling setpoint to 23 \textcelsius~ for each zone.

Despite its simplicity, the default controller manages to keep the indoor temperature inside the defined comfort range $[16, 27]$\textcelsius during a full-year evaluation period, as depicted in \autoref{fig:temperatures_default}. As expected, since this is a static and fully reactive control the energy consumption is higher than that of a consumption-aware controller, di cussed in \autoref{subsec:discussion}.

\subsection{Example~2: Rule-based controller}
\label{subsec:example2}
After testing the default controller, a simple RBC was implemented for the same environment. The input of the RBC is the averaged indoor temperature of the building zones, obtained from the observation space. If this value is above or under the comfort range, the controller respectively increases or decreases the setpoints by one degree (\textcelsius). The implementation of this controller is available in the \sinergym{} class \texttt{RBCIncrementalDatacenter}, included in the \texttt{sinergym.utils.controllers} module. This code can be extended to implement other rules, or be adapted to other environments with different observations and actions.

The RBC controller was also able to keep the indoor temperature within the comfort range in the evaluation period, as shown in \autoref{fig:temperatures_rbc}, offering a similar performance to the default controller in terms of comfort and energy consumption. Moreover, in contrast to the default controller, RBC does not require a direct modification of the building model, as its implementation is external and independent. This provides a more user-friendly reactive control alternative that is adaptable to complex scenarios.

\subsection{Example~3: DRL controllers}
\label{subsec:example3}
We evaluated three DRL algorithms: Soft Actor Critic (SAC) \cite{haarnoja2018}, Twin Delayed Deep Deterministic Policy Gradient (TD3) \cite{fujimoto2018}, and Proximal Policy Optimization (PPO) \cite{schulman2017ppo}. Their implementation was based in the StableBaselines3 library, which can be easily used with \sinergym{}. Each agent was trained during 40 episodes, each one corresponding to a 1-year simulation.

\begin{figure}[htpb]
    \centering
    \includegraphics[width=\linewidth]{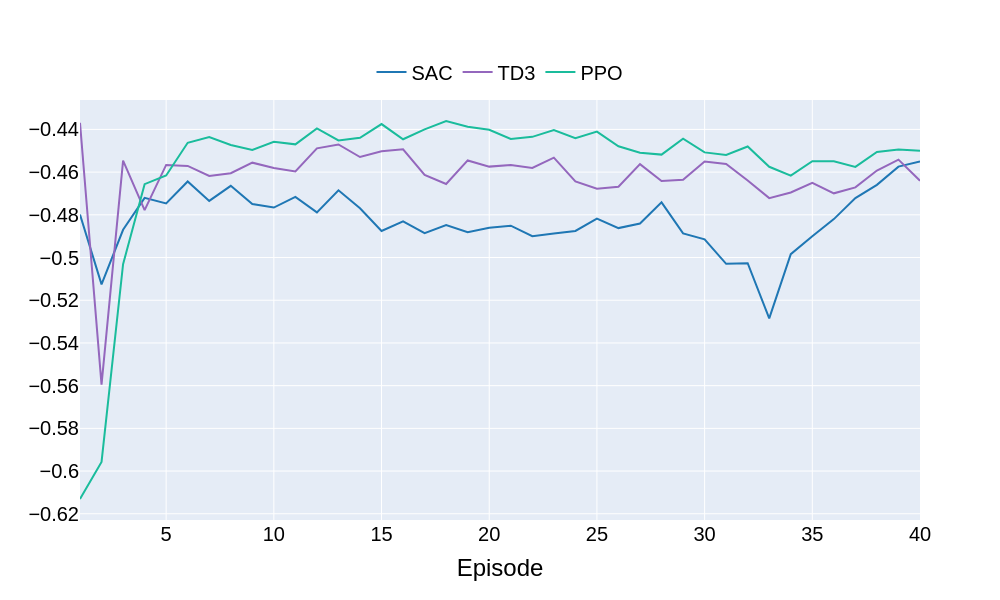}
    \caption{Mean reward of DRL agents uring training}
    \label{fig:mean_reward_progress_PPO_train}
\end{figure}

Using \sinergym{}‘s feature for intermediate evaluations, we tested the agent’s performance every 5 training episodes. This enabled us to monitor the progression of reward and BEO metrics during learning, as well as save the best-performing model obtained throughout the process. As depicted in \autoref{fig:mean_reward_progress_PPO_train}, the reward generally started stabilizing after episode 7 for all the algorithms. While PPO began as the weakest model, it quickly surpassed the others and consistently delivered better results. Overall, PPO proved to be the most stable learning algorithm in this case, followed by TD3 and SAC.

\autoref{fig:temperatures_sac}, \autoref{fig:temperatures_td3} and \autoref{fig:temperatures_ppo} show that the behavior of the DRL agents notably differed from the default controller and the RBC, allowing temperatures closer to the comfort threshold. The underlying control strategy that they autonomously learned focused on maximizing energy savings while avoiding penalties due to temperature deviations. In this scenario, the penalties for temperature and energy consumption are balanced; adjusting the reward function weights could prioritize one term over the other, generating new policies automatically without expert knowledge. Although the hyperparameters have not been optimized yet, \autoref{fig:energy_savings_per_month} show that DRL achieved energy savings in the range of 15-18\% during several months when compared to the reactive control alternatives. This will be discussed in \autoref{subsec:discussion}.

\subsection{Hyperparameter optimization}\label{subsec:fine-tuning}
Finally, we selected the controller trained with PPO, as it proved to be the most stable, as a baseline for further optimization. We initiated a Weights\&Biases sweep to explore a grid of 108 hyperparameter combinations (\autoref{tab:drl_hyperparameters}), executed in parallel on Google Cloud using the containerized version of \sinergym{}. Comprehensive training and output data—including weights, logs, final states, hyperparameters, actions, observations, rewards, and BEO metrics—were recorded in Weights\&Biases\footnote{\url{https://wandb.ai/sail_ugr/sinergym_paper}}. The performance of the best PPO configuration is illustrated in \autoref{fig:temperatures_ppo_tuned} ---the detailed results of the top 10 PPO configurations are included in \autoref{tab:best_models}. Notably, this process could also be applied to optimize other parameters not directly tied to the algorithm, such as the interval between control actions, the reward function, or the selection of observed or action variables.

\subsection{Discussion}\label{subsec:discussion}

\begin{figure}[htpb]
    \centering
    \includegraphics[width=\textwidth]{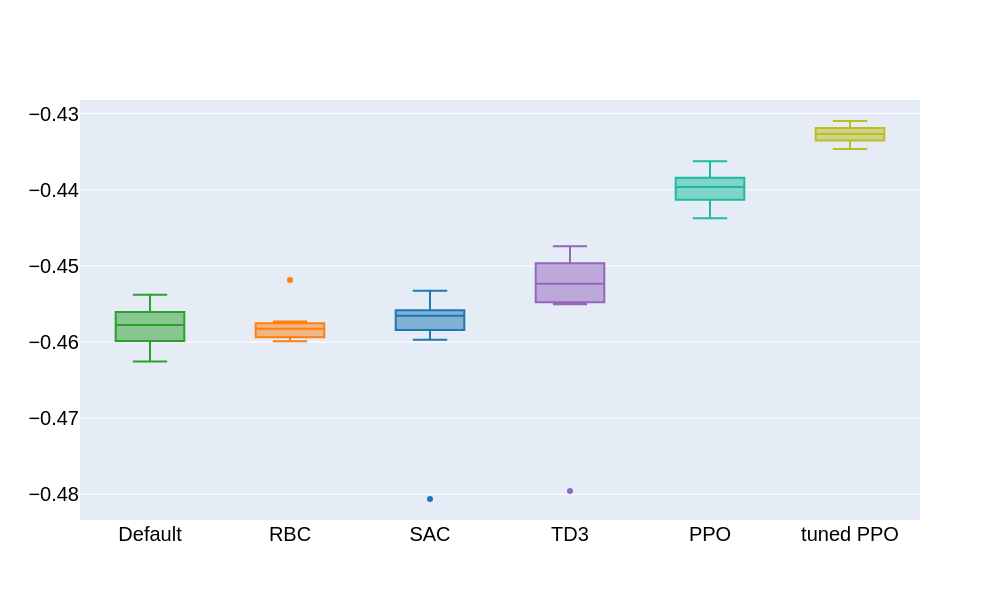}
    \caption{Mean reward obtained by Default, RBC, SAC, TD3, PPO and optimized PPO controllers during evaluation (10 episodes)}
    \label{fig:mean_reward_benchmark}
\end{figure}

The controllers were compared in terms of rewards for 10 episodes after training finished. \autoref{fig:mean_reward_benchmark} represents the rewards obtained by each one. As previously mentioned, PPO achieved the best and most consistent results. The default controller and the RBC showed similar results, with RBC being slightly more stable against different conditions. On the other hand, SAC performed similar to RBC and the default controller, while TD3 generally achieved a better performance than both of them.

\begin{figure}[htpb]
    \centering
     \includegraphics[width=\textwidth]{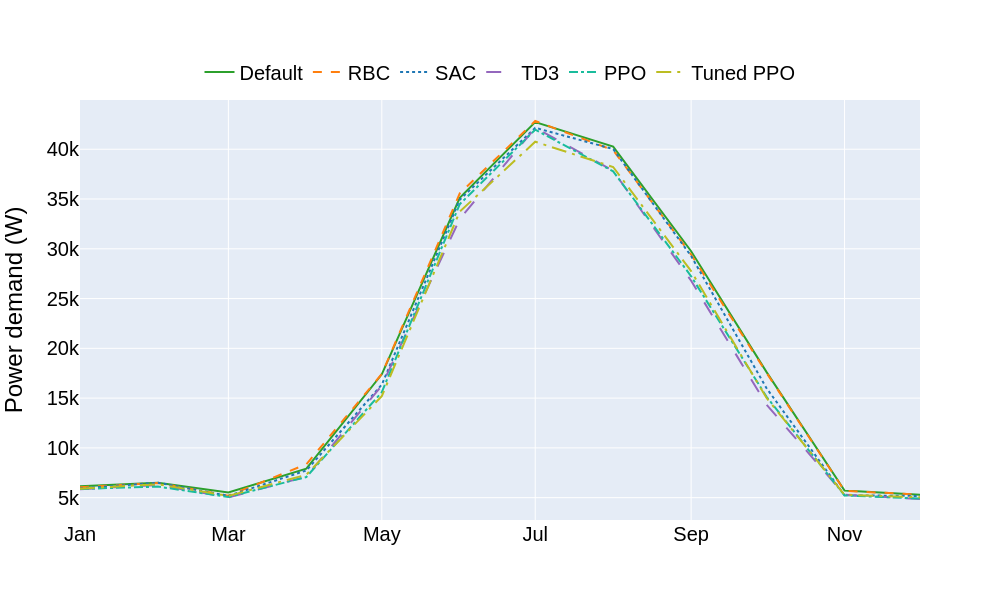}
     \caption{Monthly average power demand for each controller}
     \label{fig:energy_demand_month}
\end{figure}

\begin{figure}[htpb]
     \centering
     \begin{subfigure}[b]{\textwidth}
         \centering
         \includegraphics[width=\textwidth]{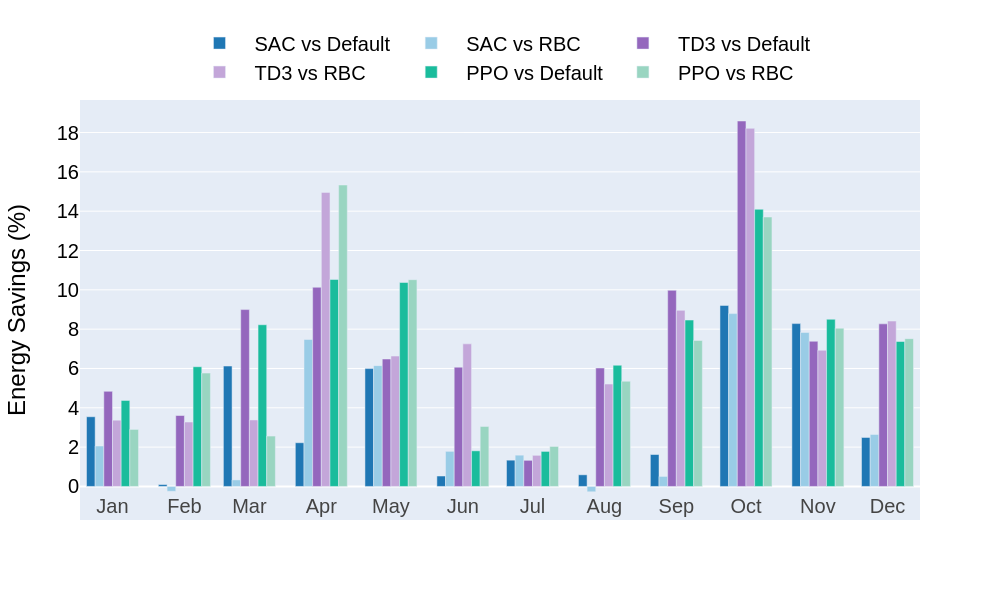}
         \caption{Energy savings (\%): comparison between DRL algorithms versus Default controller and RBC}
         \label{fig:energy_savings_per_month}
     \end{subfigure}
     \begin{subfigure}[b]{\textwidth}
         \centering
        \includegraphics[width=\linewidth]{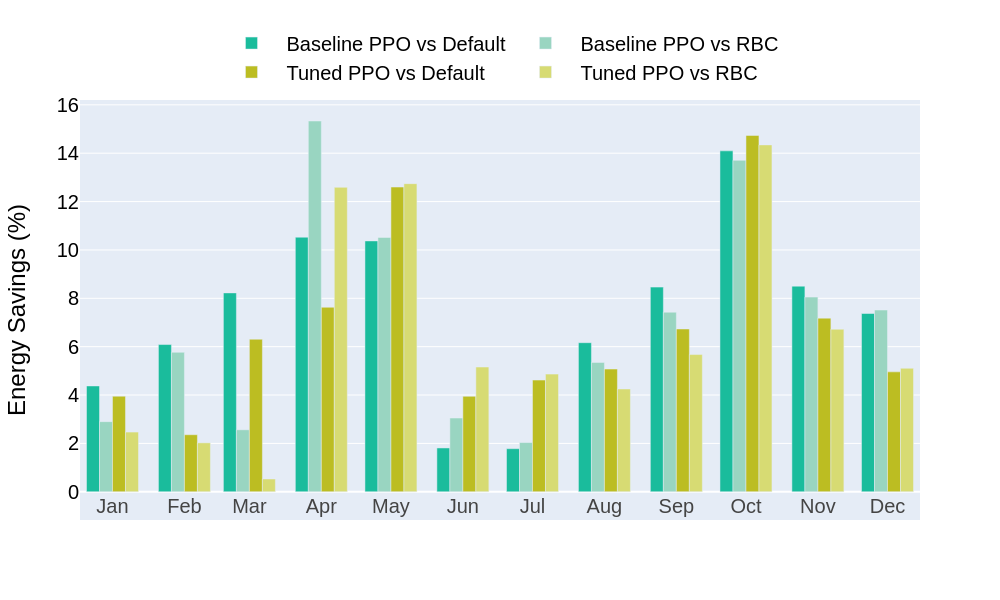}
        \caption{Energy savings (\%): comparison between the baseline PPO and the optimized (tuned) PPO versus Default controller and RBC}
        \label{fig:energy_savings_tuned}
     \end{subfigure}     
     \hfill
     
    \caption{Energy savings achieved by each DRL controller during evaluation. Darker bars represent DRL versus the default controller, while lighter bars represent DRL versus RBC}
    \label{fig:all_experiments_energy}
\end{figure}

Since the reward function combines both the comfort and the energy terms, it is interesting to analyze the values of each one separately. Accordingly, indoor temperatures are presented in \autoref{fig:all_experiments_temp} and energy demand is depicted in \autoref{fig:energy_demand_month} and \autoref{fig:all_experiments_energy}. 

The demand values represented in \autoref{fig:energy_demand_month} do not show major differences among the controllers. However, if we express them as percentages, the results in \autoref{fig:energy_savings_per_month} reveal the energy savings obtained by the DRL agents with respect to the default and the RBC controllers. Savings were mostly achieved during the intermediate seasons, from March to May and from September to November, when the temperatures are mild and the HVAC operation offers more opportunities for optimization. Also, TD3 achieved remarkable energy savings in March, September, October, and December, significantly outperforming PPO. However, PPO generally demonstrated better control over the building's internal temperature, as well as greater stability against induced climate perturbations. The optimized PPO, in turn, obtained significant total savings with respect to the default and the RBC controllers, as shown in \autoref{fig:energy_savings_tuned}.

As a side observation, we found that the percentage of time an agent drives indoor temperature outside the comfort zone is not a consistently reliable metric for evaluating controller performance. It is often better to allow temperatures to remain slightly outside the comfort zone for a bit longer, rather than having brief but significant violations. In contrast, the average temperature violation metric provided by \sinergym{} offers a more accurate assessment of the agent’s deviation from the comfort range and delivers a clearer picture of its overall performance.


\section{Conclusions}\label{sec:conclusions}
Developing and testing energy control strategies through simulation is becoming the norm in the building operation sector, as it reduces real-world experimentation risks and implementation costs. Extensive simulation is essential in RL, as agents require massive amounts of data to train effectively. As the sector advances toward intelligent, data-driven operations through digital twin technology, the demand for robust simulation platforms continues to grow. While a few existing frameworks leverage building simulation engines to support RL training, most are still in early development or lack the core functionalities necessary for integration into these advanced systems.

In this paper, we have presented \sinergym{}, a mature building simulation and control framework for RL-based BEO. It offers a scalable, customizable, and robust platform for configuring, running, and tracking massive building simulations, including numerous valuable features for RL. Specifically, the framework provides custom rewards, controllers, wrappers, continuous and discrete action spaces, support for existing RL libraries, and experiment monitoring and visualization. Moreover, \sinergym{} includes several building models that can be adapted and configured by the user, yielding tens of pre-defined environments. \sinergym{} represents a significant advancement in the field of twin-enabled building operations and control, offering remarkable ease of use, comprehensive functionalities, robust support, and detailed documentation.

The experiments presented in the paper demonstrate the capabilities of \sinergym{}'s. In particular, we have showcased the flexibility of the tool in a variety of experimental scenarios, including training and comparison of multiple controllers. The results demonstrate how DRL combined with hyperparameter optimization can lead to robust controllers that do not rely on expert knowledge. This highlights the flexibility and effectiveness of DRL in adapting to complex control problems for BEO, as well as \sinergym{}'s capabilities to automate the configuration and learning processes.

Future work in the short term will be focused on continuing the development of \sinergym{} with new features and providing support for the current functionalities. For instance, we will develop a graphical interface to facilitate the configuration and visualization of the experiments. We also plan to add support for additional co-simulation engines, starting from Modelica. The integration of \sinergym{} with CityLearn’s grid control is also a promising area for future exploration, potentially enhancing coordination between building energy and grid optimization.

Similarly to the widely used Atari and MuJoCo DRL benchmarks, our medium-term goal is to create a consistent benchmark of building environments. This will provide a set of experimental scenarios large enough to support the comprehensive and standardised comparison of different control strategies, thereby supporting future advancements in BEO.

\section*{Author's contributions}
\textbf{Alejandro Campoy-Nieves}: Conceptualization, Software, Writing -- original draft, Visualization, Validation. \textbf{Antonio Manjavacas}: Conceptualization, Software, Writing -- review and editing. \textbf{Javier Jiménez-Raboso}: Conceptualization, Software, Writing -- review and editing. \textbf{Miguel Molina-Solana}: Conceptualization, Methodology, Writing -- review and editing, Funding acquisition, Supervision. \textbf{Juan Gómez-Romero}: Conceptualization, Methodology, Writing -- review and editing, Funding acquisition, Supervision.

\section*{Declaration of interest}
The authors declare that they have no known competing financial or non-financial interests that could have appeared to influence the work reported in this paper.

\section*{Data availability}
Data and software are available at the project repository: \url{https://github.com/ugr-sail/sinergym}.

\section*{Acknowledgements}
This work was funded by FEDER/Junta de Andalucía (D3S project, P21.00247), Spanish Ministry of Science (SPEEDY, TED2021.130454B.I00) and the NextGenerationEU funds (IA4TES project, MIA.2021.M04.0008). A. Manjavacas is funded by FEDER/Junta de Andalucía under SE21\_UGR\_IFMIF-DONES project.

\appendix

\section{Settings}\label{appendix:examples_details}

This appendix provides details about the configuration of the examples detailed in \autoref{sec:examples}. In all cases, control actions are performed every 15 minutes ---i.e., 4 timesteps per hour--- and the episode's run period corresponds to a complete year (from January 1st to December 31st).

\paragraph{Observation space}
\autoref{table:observation_space} presents the variables returned by the environment as observations. 

\begin{table}[H]
\caption{Observed variables in \datacenter{}}
\centering
\label{table:observation_space}
\renewcommand{\arraystretch}{1.5} 
\begin{tabular}{lll}
\hline
\textbf{Variable name}                        & \textbf{Min.} & \textbf{Max.} \\ \hline
Site Outdoor Air Drybulb Temperature          & $-5 \cdot 10^{6}$  & $5 \cdot 10^{6}$   \\
Site Outdoor Air Relative Humidity            & $-5 \cdot 10^{6}$  & $5 \cdot 10^{6}$   \\
Site Wind Speed                               & $-5 \cdot 10^{6}$  & $5 \cdot 10^{6}$   \\
Site Wind Direction                           & $-5 \cdot 10^{6}$  & $5 \cdot 10^{6}$   \\
Site Diffuse Solar Radiation Rate per Area    & $-5 \cdot 10^{6}$  & $5 \cdot 10^{6}$   \\
Site Direct Solar Radiation Rate per Area     & $-5 \cdot 10^{6}$  & $5 \cdot 10^{6}$   \\
Zone Thermostat Heating Setpoint Temperature  & $-5 \cdot 10^{6}$  & $5 \cdot 10^{6}$   \\
Zone Thermostat Cooling Setpoint Temperature  & $-5 \cdot 10^{6}$  & $5 \cdot 10^{6}$   \\
Zone Air Temperature                          & $-5 \cdot 10^{6}$  & $5 \cdot 10^{6}$   \\
Zone Air Relative Humidity                    & $-5 \cdot 10^{6}$  & $5 \cdot 10^{6}$   \\
Zone Thermal Comfort Mean Radiant Temperature & $-5 \cdot 10^{6}$  & $5 \cdot 10^{6}$   \\
Zone Thermal Comfort Clothing Value           & $-5 \cdot 10^{6}$  & $5 \cdot 10^{6}$   \\
Zone Thermal Comfort Fanger Model PPD         & $-5 \cdot 10^{6}$  & $5 \cdot 10^{6}$   \\
Zone People Occupant Count                    & $-5 \cdot 10^{6}$  & $5 \cdot 10^{6}$   \\
People Air Temperature                        & $-5 \cdot 10^{6}$  & $5 \cdot 10^{6}$   \\
Facility Total HVAC Electricity Demand Rate   & $-5 \cdot 10^{6}$  & $5 \cdot 10^{6}$   \\
Current month                                 & 1            & 12           \\
Current day                                   & 1            & 31           \\
Current hour                                  & 0            & 23           \\ \hline
\end{tabular}
\end{table}

\paragraph{Action space}\label{subapp:action_space}
\autoref{table:action_space} presents the action space in the \datacenter{} building. Both variables are used to control both building zones simultaneously.

\begin{table}[H]
\caption{Control actions in \datacenter{}}
\label{table:action_space}
\centering
\renewcommand{\arraystretch}{1.5} 
\begin{tabular}{llll}
\hline
\textbf{Variable name} & \textbf{Default scheduler name} & \textbf{Min} & \textbf{Max} \\ \hline
Heating Setpoint RL  & Heating Setpoints       & 15.0         & 22.0         \\
Cooling Setpoint RL  & Cooling Setpoints       & 22.0         & 30.0         \\ \hline
\end{tabular}
\end{table}

\paragraph{Reward function}\label{subapp:reward_function}
The linear reward function, illustrated in \autoref{eq:linear-reward}, was used to evaluate the performance of the controllers with the parameters defined in \autoref{table:reward_function}. The comfort range was defined according to the ASHRAE standard \cite{ashrae2016}, which recommends $[18, 27]$ \textcelsius~for data centers.

\begin{table}[H]
\caption{Linear Reward function configuration}
\label{table:reward_function}
\renewcommand{\arraystretch}{1.5}
\begin{adjustbox}{width=\textwidth}
\begin{tabular}{lll}
\hline
\textbf{Parameter} & \textbf{Description}                                                                                                                  & \textbf{Value}                                                                     \\ \hline
\texttt{temperature\_variables}  & \begin{tabular}[c]{@{}l@{}}List of observed variables which are used\\ to calculate comfort metrics.\end{tabular}               & \begin{tabular}[c]{@{}l@{}}West and east \\ zone air \\ temperatures\end{tabular}            \\ \hline
\texttt{energy\_variables}      & \begin{tabular}[c]{@{}l@{}}List of the observed variables representing \\ building power demand.\end{tabular}                  & \begin{tabular}[c]{@{}l@{}}\texttt{HVAC\_}\\ \texttt{electricity\_}\\ \texttt{demand\_rate}\end{tabular} \\ \hline
\texttt{range\_comfort\_winter}  & \begin{tabular}[c]{@{}l@{}}Temperature comfort range for cold season. \\ Depends on the problem definition.\end{tabular}              & (18.0, 27.0) \textcelsius                                                                       \\ \hline
\texttt{range\_comfort\_summer}  & \begin{tabular}[c]{@{}l@{}}Temperature comfort range for hot season. \\ Depends on the problem definition.\end{tabular}               & (18.0, 27.0) \textcelsius                                                                      \\ \hline
\texttt{summer\_start}           & Month and day when summer season starts.                                                                                              & (6, 1)                                                                             \\ \hline
\texttt{summer\_final}           & Month and day when summer season ends.                                                                                                & (9, 30)                                                                            \\ \hline
\texttt{energy\_weight}          & \begin{tabular}[c]{@{}l@{}}Weight for energy reward component. Comfort \\ reward component will be $1 - W_{energy}$\end{tabular}      & 0.5                                                                                \\ \hline
\texttt{lambda\_energy}          & \begin{tabular}[c]{@{}l@{}}Constant to calibrate the energy reward component \\ magnitude. Depends on the building.\end{tabular} & 0.00005                                                                             \\ \hline
\texttt{lambda\_temperature}     & \begin{tabular}[c]{@{}l@{}}Constant to calibrate the energy reward component \\ magnitude. Depends on the building.\end{tabular} & 1.0                                                                                \\ \hline
\end{tabular}
\end{adjustbox}
\end{table}

\paragraph{Wrappers}\label{subapp:applied_wrappers}
The data logging wrapper (\texttt{LoggerWrapper}) and the 
normalization wrappers for the action (\texttt{NormalizeAction}) and the observation  (\texttt{NormalizeObservation}) spaces were applied. The \texttt{CSVLogger} and the \texttt{WandBLogger} wrappers were enabled to register output data in CSV files and Weights\&Biases.

\paragraph{RL algorithm configuration and hyperparameters}\label{subapp:hyperparameters}
The hyperparameters and neural network architecture used to train the DRL agents in \autoref{subsec:discussion} follow the default configurations of StableBaselines3. Specifically, the neural network consists of two fully connected layers with 64 units per layer for PPO, 256 units per layer for SAC, and 400 and 300 units respectively for each layer in TD3.

\autoref{tab:drl_hyperparameters} presents the grid search in the hyperparameter optimization developed with PPO algorithm in \autoref{subsec:fine-tuning}. Each experiment was executed during 20 episodes, with intermediate evaluations every 3 episodes.

\begin{table}[htpb]
\caption{Hyperparameter space used for PPO grid search optimization (108 total combinations)}
\centering
\label{tab:drl_hyperparameters}
\renewcommand{\arraystretch}{1.5} 
\begin{tabular}{lll}
\hline
\textbf{Hyperparameter}   & \textbf{Values}    & \textbf{Description} \\ \hline
learning\_rate            & 0.0001, 0.0003, 0.001     & Adjusts the step size for model updates. \\
gamma                     & 0.9, 0.99, 0.999          & Discount factor for future rewards. \\
batch\_size               & 64, 128                   & Number of samples per training step. \\
max\_grad\_norm           & 0.5, 0.9                  & Clips gradients to stabilize training. \\
ent\_coef                 & 0, 0.001, 0.01            & Controls the balance between exploration and exploitation. \\ \hline
\end{tabular}
\end{table}

\section{Hyperparameter optimization result}
\autoref{tab:best_models} shows the performance metrics and the configuration of the top 10 controllers obtained after optimizing PPOs's hyperparameters.

\begin{table}[htpb]
\caption{Top 10 PPO agents based on the best mean reward obtained during on-training evaluations. Values rounded to 4 decimals.}
\centering
\renewcommand{\arraystretch}{1.5}
\begin{tabular}{llllllllll}
\toprule
\textbf{PPO} & \textbf{Best} & \textbf{Temp.} & \textbf{Time} & \textbf{Power} & \textbf{Learn.} & \textbf{Gam.} & \textbf{Batch} & \textbf{Max.} & \textbf{Ent.} \\
             \textbf{Model} & \textbf{reward} & \textbf{dev. (\textcelsius)} & \textbf{dev. (\%)} & \textbf{dem. (W)} & \textbf{rate} &              & \textbf{size} & \textbf{norm.} & \textbf{coef.} \\
\midrule
\textbf{1st}  & -0.4310  & 0.0009  & 0.9389  & 17221.1037  & 0.0003  & 0.9  & 128  & 0.9  & 0 \\
\textbf{2nd}  & -0.4318  & 0.0059  & 5.4366  & 17152.8005  & 0.0003  & 0.9  & 64   & 0.5  & 0 \\
\textbf{3rd}  & -0.4336  & 0.0003  & 0.4937  & 17338.2957  & 0.0003  & 0.9  & 128  & 0.9  & 0.01 \\
\textbf{4th}  & -0.4342  & 0.0017  & 1.5097  & 17331.0688  & 0.0003  & 0.9  & 64   & 0.9  & 0.01 \\
\textbf{5th}  & -0.4346  & 0.0013  & 1.4668  & 17355.1309  & 0.0003  & 0.9  & 64   & 0.5  & 0.01 \\
\textbf{6th}  & -0.4347  & 0.0063  & 5.0884  & 17262.0097  & 0.0003  & 0.9  & 64   & 0.9  & 0 \\
\textbf{7th}  & -0.4348  & 0.0006  & 0.3881  & 17380.3400  & 0.0003  & 0.9  & 128  & 0.9  & 0.001 \\
\textbf{8th}  & -0.4348  & 0.0027  & 2.2716  & 17338.9685  & 0.0003  & 0.9  & 64   & 0.9  & 0.001 \\
\textbf{9th}  & -0.4351  & 0.0011  & 0.6150  & 17379.9875  & 0.001   & 0.9  & 128  & 0.5  & 0.001 \\
\textbf{10th} & -0.4356  & 0.0035  & 2.3630  & 17352.6699  & 0.0003  & 0.9  & 64   & 0.5  & 0.001 \\ \hline \hline
\textbf{baseline}  & -0.4377  & 0.0043  & 2.0134  & 17423.4776  & 0.0003  & 0.99  & 64  & 0.5  & 0 \\
\hline
\end{tabular}
\subcaption*{\textbf{Best reward}: Best mean reward obtained during on-training evaluations. \textbf{Temp. dev. (\textcelsius)}: Mean deviation in degrees from comfort range.
\textbf{Time dev. (\%)}: Percentage of simulation time when air temperature deviates from the comfort range. \textbf{Power dem. (W)}: Mean Power demand.
\textbf{Learn. rate}: Learning rate. \textbf{Gam.}: Gamma discount factor. \textbf{Batch size}: Number of samples. \textbf{Max. norm.}: Maximum gradient norm clipping. \textbf{Ent. coef.}: Entropy coefficient.}
\label{tab:best_models}
\end{table}

\bibliographystyle{elsarticle-num} 
\bibliography{bibliography}

\end{document}